\documentclass[utf8]{article}
\usepackage{authblk}
\usepackage{amsmath,amsfonts}
\usepackage{algorithmic}
\usepackage{algorithm}
\usepackage{xcolor}
\usepackage{array}
\usepackage{graphicx}
\usepackage{textcomp}
\usepackage{float}
\usepackage{stfloats}
\usepackage{url}
\usepackage{verbatim}
\usepackage{tabularx}
\usepackage{multirow}
\usepackage{makecell}
\usepackage{pdfcomment}
\usepackage{hyperref}
\usepackage{lineno}
\usepackage{microtype}
\usepackage[onehalfspacing]{setspace}
\usepackage{natbib}
\usepackage{pdfpages}
\usepackage[left=2cm,right=2cm,top=2cm,bottom=2cm]{geometry}

\newcolumntype{C}{>{\centering\arraybackslash}X}

\DeclareMathOperator*{\argmin}{\arg\!\min}

\begin{document}
\onecolumn

\title{Paired Competing Neurons Improving STDP Supervised Local Learning In Spiking Neural Networks}
\author[1]{Gaspard Goupy}
\author[1]{Pierre Tirilly}
\author[1]{Ioan Marius BILASCO}
\affil[1]{Univ. Lille, CNRS, Centrale Lille, UMR 9189 CRIStAL, F-59000 Lille, France}
\date{}

\maketitle

\begin{abstract}
    Direct training of Spiking Neural Networks (SNNs) on neuromorphic hardware has the potential to significantly reduce the energy consumption of artificial neural network training.
    SNNs trained with Spike Timing-Dependent Plasticity (STDP) benefit from gradient-free and unsupervised local learning, which can be easily implemented on ultra-low-power neuromorphic hardware.
    However, classification tasks cannot be performed solely with unsupervised STDP.
    In this paper, we propose Stabilized Supervised STDP (S2-STDP), a supervised STDP learning rule to train the classification layer of an SNN equipped with unsupervised STDP for feature extraction.
    S2-STDP integrates error-modulated weight updates that align neuron spikes with desired timestamps derived from the average firing time within the layer.
    Then, we introduce a training architecture called Paired Competing Neurons (PCN) to further enhance the learning capabilities of our classification layer trained with S2-STDP.
    PCN associates each class with paired neurons and encourages neuron specialization toward target or non-target samples through intra-class competition.
    We evaluate our methods on image recognition datasets, including MNIST, Fashion-MNIST, and CIFAR-10.
    Results show that our methods outperform state-of-the-art supervised STDP learning rules, for comparable architectures and numbers of neurons.
    Further analysis demonstrates that the use of PCN enhances the performance of S2-STDP, regardless of the hyperparameter set and without introducing any additional hyperparameters.

\end{abstract}

\section{Introduction}
Artificial Neural Networks (ANNs) have gathered exponential attention across diverse domains in recent years~\citep{abiodunStateoftheartArtificialNeural2018}.
However, ANN training suffers from high and inefficient energy consumption on modern computers based on the von Neumann architecture~\citep{zouBreakingNeumannBottleneck2021}.
Spiking Neural Networks (SNNs)~\citep{ponulakIntroductionSpikingNeural2011} implemented on neuromorphic hardware~\citep{schumanSurveyNeuromorphicComputing2017,shresthaSurveyNeuromorphicComputing2022} have emerged as a promising solution to overcome the von Neumann bottleneck~\citep{zouBreakingNeumannBottleneck2021} and enable energy-efficient computing.
In particular, memristive-based neuromorphic hardware~\citep{jeongMemristorsEnergyEfficientNew2016,xuAdvancesMemristorBasedNeural2021} is an excellent candidate for ultra-low-power applications, potentially reducing energy consumption by at least one order of magnitude compared to state-of-the-art CMOS-based neuromorphic hardware~\citep{miloMemristiveCMOSDevices2020,liuLowPowerComputingNeuromorphic2021}, and by several orders of magnitude compared to GPUs~\citep{yaoFullyHardwareimplementedMemristor2020,liSituLearningHardware2022}.
However, direct training of SNNs on neuromorphic hardware comes with a major constraint: implementing network-level communication is challenging and requires significant circuitry overhead~\citep{zenkeBrainInspiredLearningNeuromorphic2021}.
Therefore, the involved learning mechanisms should rely on local weight updates, i.e., updates based only on the two neurons that the synapse connects.

Training SNNs to achieve state-of-the-art performance is typically accomplished with adaptations of Backpropagation (BP)~\citep{eshraghianTrainingSpikingNeural2021,dampfhofferBackpropagationBasedLearningTechniques2023}.
However, these methods are challenging to implement on neuromorphic hardware as they rely on non-local learning.
In addition, they employ gradient approximation to circumvent the non-differentiable nature of the spike generation function, which is suboptimal.
Other approaches attempted to make gradient computation local, notably by utilizing feedback connections~\citep{neftciEventDrivenRandomBackPropagation2017,zenkeSuperSpikeSupervisedLearning2018}, or by employing a layer-wise cost function~\citep{maTemporalDependentLocal2021,mirsadeghiSTiDiBPSpikeTime2021}.
Yet, they do not solve the gradient approximation problem.
Furthermore, all of the aforementioned BP-based methods rely solely on supervised learning, which increases the dependence on labeled data.
We believe that machine learning algorithms should minimize this dependence on supervision by employing unsupervised feature learning~\citep{bengioRepresentationLearningReview2013}.
Hence, an optimal classification system should include both unsupervised and supervised components, for data representation and classification, respectively.

Spike Timing-Dependent Plasticity (STDP)~\citep{caporaleSpikeTimingDependent2008} is a gradient-free, unsupervised and local alternative to BP, inspired by the principal form of plasticity observed in biological synapses~\citep{hebbOrganizationBehavior1949}.
STDP solves the previously mentioned limitations of BP and is inherently implemented in memristor circuits~\citep{querliozSimulationMemristorbasedSpiking2011,schumanSurveyNeuromorphicComputing2017}, which makes it suitable for on-chip training on memristive-based neuromorphic hardware~\citep{saighiPlasticityMemristiveDevices2015,khacefSpikebasedLocalSynaptic2023}.
Unsupervised feature learning with STDP has been extensively studied in the literature, particularly for image recognition tasks.
Convolutional SNNs (CSNNs) trained with STDP have demonstrated the ability to improve data representation by extracting relevant features from images~\citep{tavanaeiMultilayerUnsupervisedLearning2017,ferreUnsupervisedFeatureLearning2018,kheradpishehSTDPbasedSpikingDeep2018,falezMultilayeredSpikingNeural2019,srinivasanReStoCNetResidualStochastic2019}.
However, to perform classification based on the extracted features, these solutions employ external classifiers, such as ANNs or support vector machines, which are incompatible with neuromorphic hardware.
To leverage the potential of these CSNNs in neuromorphic hardware and enable end-to-end SNN solutions, spiking classifiers trained with local supervised learning rules must be designed.
Ensuring compatibility between classifiers and CSNNs, particularly regarding the type of local rule employed, could significantly mitigate hardware implementation overhead.

Although STDP is traditionally formulated for unsupervised learning, it can be adapted for supervised learning by incorporating a third factor, taking the form of an error signal that is used to guide the STDP updates~\citep{fremauxNeuromodulatedSpikeTimingDependentPlasticity2015}.
As a result, STDP enables end-to-end SNNs to perform classification tasks by combining unsupervised STDP for feature extraction and supervised STDP for classification~\citep{shresthaStableSpiketimingDependent2017,mozafariBioinspiredDigitRecognition2019,thieleEventBasedTimescaleInvariant2018,leeDeepSpikingConvolutional2019}.
Several supervised adaptations of STDP are reported in the literature~\citep{ponulakSupervisedLearningSpiking2010,shresthaStableSpiketimingDependent2017,tavanaeiBPSTDPApproximatingBackpropagation2019,shresthaApproximatingBackpropagationBiologically2019,leeDeepSpikingConvolutional2019,haoBiologicallyPlausibleSupervised2020,zhaoGLSNNMultiLayerSpiking2020,zhangTuningConvolutionalSpiking2021,saraniradAssemblybasedSTDPNew2022}.
Yet, all of the aforementioned rules are designed to train SNNs with multiple spikes per neuron, which is undesirable because state-of-the-art CSNNs trained with unsupervised STDP usually employ one spike per neuron.
For compatibility with these CSNNs, a spiking classifier trained with supervised STDP should adhere to this single-spike approach.
In addition, it has been shown that using one spike per neuron with temporal coding presents several advantages for visual tasks, including fast information transfer, low computational cost, and improved energy efficiency~\citep{rullenRateCodingTemporal2001,parkT2FSNNDeepSpiking2020,guoNeuralCodingSpiking2021}.
The literature exploring supervised adaptations of STDP for training SNNs with only one spike per neuron is limited.
Reward-modulated STDP (R-STDP)~\citep{mozafariBioinspiredDigitRecognition2019} is a learning rule based on Winner-Takes-All (WTA) competition~\citep{ferreUnsupervisedFeatureLearning2018} that modulates the polarity of the STDP update to apply a reward or a punishment.
R-STDP has gained popularity notably for its simplicity, but it results in inaccurate weight updates as only the polarity of STDP is adjusted.
Recently, Supervised STDP (SSTDP)~\citep{liuSSTDPSupervisedSpike2021} proposes a method to modulate, in the output layer, both the polarity and intensity of STDP with temporal errors, resulting in more accurate weight updates.
When combined with the non-local optimization process of BP, SSTDP enables state-of-the-art performance in deep SNNs on various image recognition datasets.
However, it has not yet been investigated in settings based on local learning, combining unsupervised STDP for feature extraction and SSTDP for classification.
In addition, we claim that SSTDP faces two issues that may limit its performance.
First, SSTDP training results in a limited number of STDP updates per epoch, which can lead to premature training convergence.
Second, SSTDP training causes the saturation of firing timestamps toward the maximum firing time, which can limit the ability of the SNN to separate classes.

\begin{figure}[!ht]
    \begin{center}
        \includegraphics[width=\textwidth]{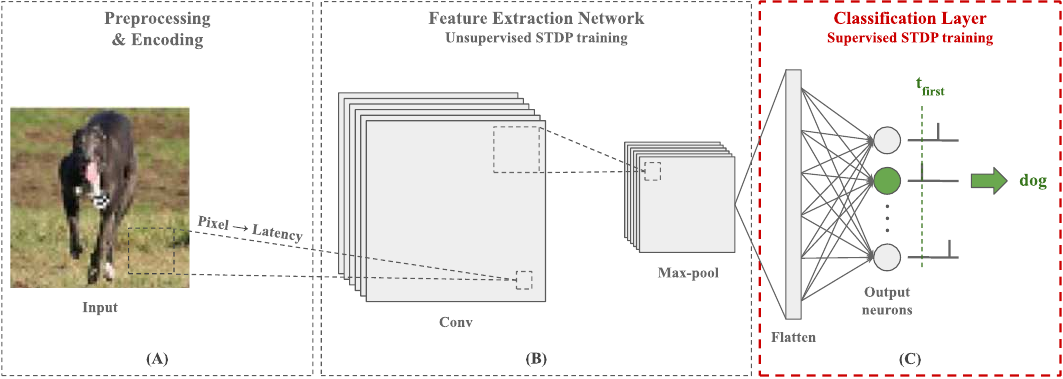}
    \end{center}
    \caption{Architecture of the SNN employed in this paper for image recognition tasks. First, the image is preprocessed and each pixel is encoded into a single floating-point timestamp using the latency coding scheme. Then, a Convolutional SNN (CSNN) trained with unsupervised STDP is used to extract relevant features from the image. The resulting feature maps are compressed through a max-pooling layer to reduce their size and provide invariance to translation on the input. Lastly, they are flattened and fed to a fully-connected SNN trained with a supervised adaptation of STDP for classification. Each output neuron is associated with a class and the first one to fire predicts the label. Training is done in a layer-wise fashion. This classification pipeline organized into 3 blocks provides a flexible framework for SNNs combining feature extraction and classification. In this paper, we focus on the classification layer block~(C), which may be integrated after other encoding or feature extraction blocks based on latency coding.}
    \label{fig:architecture}
\end{figure}
In this paper, we focus on the supervised STDP training of a spiking classification layer with one spike per neuron and temporal decision-making.
This classification layer is the output layer of an SNN equipped with unsupervised STDP for feature extraction, as illustrated in Figure~\ref{fig:architecture}.
The main contributions of this paper include the following:
\begin{enumerate}
    \item In a preliminary study, we analyze the behavior of SSTDP when used to train the classification layer of an SNN equipped with unsupervised STDP. We demonstrate that the rule encounters two issues that may limit its performance: the limited number of STDP updates per epoch and the saturation of firing timestamps toward the maximum firing time.
    \item To address the issues of SSTDP, we propose Stabilized Supervised STDP (S2-STDP), a supervised STDP learning rule that teaches neurons to align their spikes with dynamically computed desired timestamps derived from the average firing time within the layer.
    \item To further enhance the learning capabilities of our classification layer trained with S2-STDP, we introduce a training architecture called Paired Competing Neurons (PCN). This method associates each class with paired neurons and encourages neuron specialization toward target or non-target samples through intra-class competition.
    \item We evaluate the performance of S2-STDP and PCN on three image recognition datasets of growing complexity: MNIST, Fashion-MNIST, and CIFAR-10.
\end{enumerate}

The remainder of this paper is organized as follows.
In Section~\ref{sec:Background}, we provide the necessary background information about the SNN employed in this study.
In Section~\ref{sec:Problem}, we demonstrate experimentally the aforementioned issues of SSTDP, which we address with our contributions.
In Section~\ref{sec:Methods}, we describe our spiking classification layer and our proposed training methods.
In Section~\ref{sec:Results}, we cover our results on image recognition datasets and provide an in-depth investigation of the key characteristics of our methods.
In Section~\ref{sec:Conclusion}, we conclude the paper.
The source code is publicly available at: \url{https://gitlab.univ-lille.fr/fox/snn-pcn}.

\section{Background} \label{sec:Background}

\subsection{Neural coding}
Since spiking neurons communicate through spikes, encoding the image, as illustrated in Figure~\ref{fig:architecture}(A), is a necessary step before the SNN can process it.
In this work, we use a temporal coding scheme called latency coding~\citep{thorpeSpikebasedStrategiesRapid2001}.
This scheme represents each pixel by a single spike timestamp, thus limiting the number of generated spikes and making the coding more energy efficient, which is suitable for implementation on ultra-low-power devices.
For a given pixel $x \in \left[0,1\right]$, we calculate its spike timestamp $t\left(x\right)$ as follows:
\begin{equation}
    t\left(x\right)= T_{\mathrm{max}} \cdot \left(1 - x\right),
    \label{eq:latency-coding}
\end{equation}
where $T_{\mathrm{max}}$ is the maximum firing time (set to 1 in this work).
Consequently, the intensity of the pixel is encoded through latency: higher pixel values correspond to lower latencies, and vice versa.
Spike timestamps are represented by floating-point values to align with spike-based communication in event-driven neuromorphic hardware.

\subsection{Neuron model}
To simulate the dynamics of spiking neurons, we use the Single-Spike Integrate-and-Fire (SSIF) model~\citep{kheradpishehTemporalBackpropagationSpiking2020,goupyUnsupervisedEfficientLearning2023}, describing IF neurons that can fire at most once per sample.
Following latency coding, SSIF neurons encode the magnitude of their activation through spike timing: neurons firing first are the most strongly activated.
The neurons integrate input spikes to their membrane potential $V$ as follows:
\begin{equation}
    \frac{\partial V_j\left(t\right)}{\partial t} = \sum_i w_{ij} \cdot S_i\left(t\right),
    \label{eq:if-model}
\end{equation}
where $t$ is the timestamp, $V_j$ is the membrane potential of neuron $n_j$, $w_{ij}$ is the weight of the synapse from $n_i$ to $n_j$, and $S_i\left(t\right)$ indicates the presence ($S_i\left(t\right)=1$) or absence ($S_i\left(t\right)=0$) of a spike from input neuron $n_i$ at timestamp $t$.
When the membrane potential of neuron $n_j$ exceeds a defined threshold $V_{\mathrm{th}}$ (i.e. $V_j\left(t\right) \geq V_{\mathrm{th}}$), the neuron emits a spike at timestamp $t$ and is deactivated until the next sample is shown.

\subsection{Unsupervised STDP feature learning} \label{sec:CSNN}
Unsupervised training of CSNNs with STDP, as illustrated in Figure~\ref{fig:architecture}(B), is an effective approach for improving image representation before classification without using any labeled data.
In this work, we use the CSNN model introduced by~\cite{falezMultilayeredSpikingNeural2019}.
This model comprises trainable convolutional layers that extract spatial features from an encoded image, and non-trainable max-pooling layers that reduce the size of the feature maps and provide translation invariance on the input.
The output of the CSNN consists of two-dimensional feature maps containing the output spikes of the neurons in the final layer.
Following the SSIF neuron model, there is at most one spike per neuron (i.e. per feature) and the intensity of the activation is encoded temporally.

STDP training consists in adjusting synaptic weights of the convolutional layers based on the time difference between input and output neuron spikes.
The rule considers both causal and non-causal relationships, increasing the synaptic weight when the input neuron fires before the output neuron (causal), and decreasing it otherwise (non-causal).
For an output neuron $n_j$, its weights are updated as follows with the multiplicative STDP~\citep{querliozSimulationMemristorbasedSpiking2011}:
\begin{equation}
    \Delta w_{ij}=\begin{cases}
        A^+ \times \exp\left(-\beta\frac{ w_{ij}-w_{\mathrm{min}}}{w_{\mathrm{max}}-w_{\mathrm{min}}}\right) & \text{if } t_j \geq t_i \\
        A^- \times \exp\left(-\beta\frac{w_{\mathrm{max}}-w_{ij}}{w_{\mathrm{max}}-w_{\mathrm{min}}}\right)  & \text {o.w. }
    \end{cases},
    \label{eq:stdp-mult}
\end{equation}
where $w_{ij}$ represents the weight of the synapse connecting input neuron $n_i$ and output neuron $n_j$, $\Delta w_{ij}$ is the weight change, $A^+$ and $A^-$ are the positive and negative learning rates, $\beta$ is the saturation factor, $w_{\mathrm{min}}$ and $w_{\mathrm{max}}$ are the minimum and maximum achievable weight values in the layer, and $t_i$ is the firing timestamp of neuron $n_i$.
In addition to STDP, WTA competition and homeostatic plasticity are employed in the convolutional layers~\citep{falezMultilayeredSpikingNeural2019}.
Via lateral inhibition, WTA competition promotes the learning of various patterns: for each sample, only the weights of the first neuron to fire are updated.
Via threshold adaptation, homeostatic plasticity regulates the WTA competition.

We recall that we focus, in this work, on the spiking classification layer, given the extensive research already existing on CSNNs trained with unsupervised STDP.
For an in-depth explanation and investigation of the employed CSNN model, we refer readers to~\citep{falezUnsupervisedVisualFeature2019,falezMultilayeredSpikingNeural2019,falezImprovingSTDPbasedVisual2020}.
In addition, other feature extractors may be employed instead of this model, provided that the resulting output features adhere to the principles of the SSIF model.

\subsection{SSTDP} \label{seq:sstdp}
Supervised STDP (SSTDP)~\citep{liuSSTDPSupervisedSpike2021} is a supervised learning rule that combines the weight updates of STDP with the global optimization process of BP to train deep SNNs with one spike per neuron.
This rule obtains state-of-the-art performance in diverse visual tasks but it relies on non-local computation for multi-layer architectures, which is difficult to implement on neuromorphic hardware~\citep{zenkeBrainInspiredLearningNeuromorphic2021}.
Given the architecture of the SNN employed in this work (see Figure~\ref{fig:architecture}), we briefly introduce the SSTDP training process at the output layer only.

SSTDP training consists in teaching output neurons to fire within one of two desired time ranges.
For each sample, these time ranges are dynamically computed for the target neuron (i.e. the neuron associated with the class of the sample) and the non-target neurons (i.e. the neurons not associated with the class of the sample), based on the average firing time $T_{\mathrm{mean}}$ in the layer.
At the end of the presentation of a sample, each output neuron updates its weights with an error-modulated STDP.
The error $e_j$ of neuron $n_j$ is defined as the temporal difference between the neuron firing timestamp and its time range boundary, which can be formulated as:
\begin{equation}
    e_j=\begin{cases}
        \max \left\{ 0, t_j - \left(T_{\mathrm{mean}} - g_1 \right)\right\} & \text{if } c_j = y    \\
        \min \left\{ 0, t_j - \left(T_{\mathrm{mean}} + g_2 \right)\right\} & \text{if } c_j \neq y
    \end{cases},
    \label{eq:sstdp-tdesired}
\end{equation}
where $t_j$ is the firing timestamp of neuron $n_j$, $g_1$ and $g_2$ are defined time gaps that control the distance from $T_{\mathrm{mean}}$, $c_j$ is the class of neuron $n_j$, and $y$ is the class of the sample.
In other words, the time range of the target neuron ($c = y$) is $\left[0, T_{\mathrm{mean}} - g_1\right]$ and the time range of the non-target neurons ($c \neq y$) is $\left[T_{\mathrm{mean}} + g_2, T_{\mathrm{max}}\right]$.
If a neuron fires within its desired range, the error is zero and the neuron weights are not updated.
Note that, due to the $\min$ and $\max$ functions, SSTDP exclusively permits positive errors for target neurons (promoting earlier firing) and negative errors for non-target neurons (promoting later firing), potentially restricting accurate control over the output firing timestamps.

\section{Preliminary study} \label{sec:Problem}
In Section~\ref{seq:sstdp}, we briefly outlined the SSTDP training method for an output layer with one spike per neuron.
While this rule achieves state-of-the-art performance, the formulation of the desired time ranges (as presented in Equation~\ref{eq:sstdp-tdesired}) raises a concern: SSTDP offers restricted control over the output firing timestamps.
To demonstrate this, we conducted preliminary experiments on the MNIST~\citep{lecunGradientbasedLearningApplied1998} and the Fashion-MNIST~\citep{xiaoFashionMNISTNovelImage2017} datasets, using SSTDP to train the classification layer of an SNN equipped with unsupervised STDP, as illustrated in Figure~\ref{fig:architecture}.
It led us to identify two primary issues that we aim to resolve in this work:
\begin{enumerate}
    \item the limited number of STDP updates per epoch;
    \item the saturation of firing timestamps toward the maximum firing time.
\end{enumerate}

\begin{figure}[!ht]
    \centering
    \includegraphics[width=0.78\textwidth]{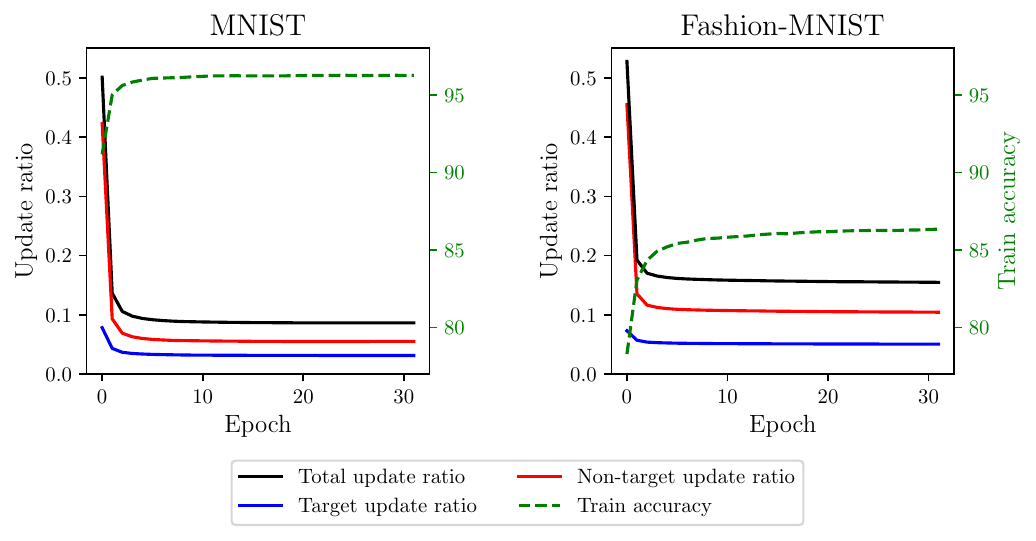}
    \caption{Update ratio and train accuracy per epoch in the classification layer trained with SSTDP. Training with SSTDP results in a limited number of STDP updates per epoch, which may lead to premature training convergence and suboptimal model performance.}
    \label{fig:sstdp_nb_updates}
\end{figure}

First, if a neuron fires within its desired time range during training, its weights are not updated.
Since neurons can easily reach their desired time range, training with SSTDP results in a limited number of updates per epoch.
Figure~\ref{fig:sstdp_nb_updates} illustrates the update ratio per epoch, computed as the average number of updates per neuron divided by the number of samples.
For both datasets, the total update ratio is around $50$\% in the initial epoch, implying that neurons do not update their weights for half of the training samples.
As the number of epochs increases, this ratio quickly decreases to $9$\% for MNIST and $16$\% for Fashion-MNIST.
This limited number of updates leads to rapid training convergence, as indicated by the stabilization of training accuracies within a few epochs.
However, since many samples are not involved in the training process, we believe that such rapid convergence is premature and may reduce the capabilities of the SNN to generalize, resulting in suboptimal model performance.
In addition, this training process is inefficient because many samples undergo computational processing by the SNN without producing weight updates.

\begin{figure}[!ht]
    \centering

    \includegraphics[width=0.78\textwidth]{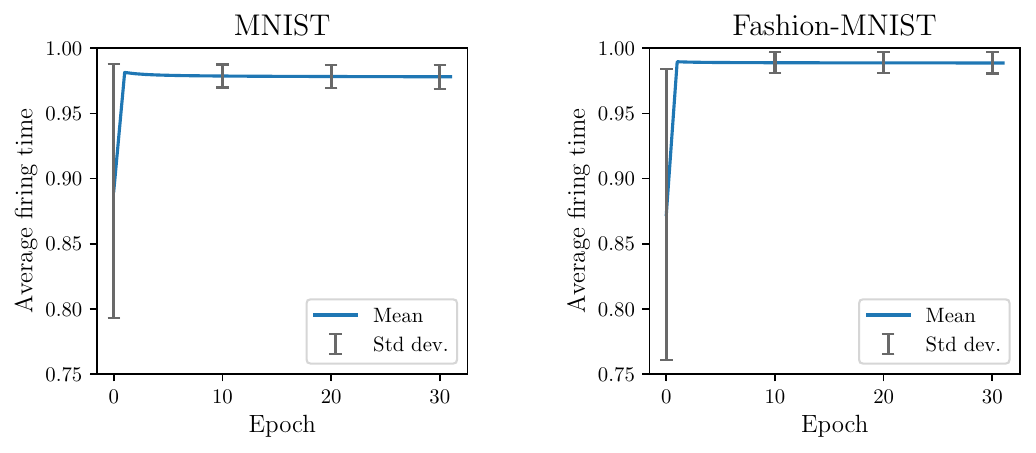}
    \caption{Average firing time per epoch in the classification layer trained with SSTDP. Training with SSTDP causes the saturation of firing timestamps toward the maximum firing time, which may limit the expressivity of the SNN and its ability to separate classes.}
    \label{fig:sstdp_tmean_evol}
\end{figure}

Second, because non-target updates are more frequent than target updates (see Figure~\ref{fig:sstdp_nb_updates}), neurons are continually pushed to fire later.
It creates a saturation effect where the firing timestamps of neurons rapidly approach the maximum firing time.
As observed in Figure~\ref{fig:sstdp_tmean_evol}, the average firing time grows rapidly in the first epochs and then stabilizes close to the maximum firing time.
During the last training epoch, we compute an average firing time of $0.98\pm0.009$ for MNIST and $0.99\pm0.008$ for Fashion-MNIST.
Note that the standard deviation corresponds to the variance in the average firing time between samples: it indicates how much the average firing time varies across samples during an epoch.
We believe that the observed saturation effect limits the expressivity of the SNN (i.e. its ability to capture discriminant spatiotemporal information) and defeats the principles of temporal coding as the majority of the input spikes are integrated by the output neurons.
In addition, the low variance in the average firing time across samples of different classes may affect the ability of the SNN to separate classes.

\section{Methods} \label{sec:Methods}
To address the issues of SSTDP, we propose S2-STDP for training a spiking classification layer with one spike per neuron and temporal decision-making.
S2-STDP employs error-modulated weight updates with dynamically computed desired timestamps derived from the average firing time within the layer.
Then, we propose the PCN training architecture, which further enhances the learning capabilities of our classification layer trained with S2-STDP.
PCN encourages neuron specialization through intra-class competition and does not introduce any additional hyperparameters.

\subsection{Supervised classification layer}
The classification layer is the output layer of our SNN, illustrated in Figure~\ref{fig:architecture}(C).
It is a fully-connected architecture composed of SSIF neurons, designed to process latency-coded spike inputs (i.e. with at most one spike per input, and where the intensity of the input is encoded temporally).
For a $N$-class problem, the classification layer comprises $N$ neurons $\left(n_1, \ldots, n_N\right)$, each associated with a distinct class $\left(c_1, \ldots, c_N\right)$.
The purpose of the classification layer is to predict the class of a given sample.
Since SSIF neurons can fire at most once per sample, we make this prediction based on the output neuron that fires first.
This method eliminates the need to propagate the entire input for inference, which can reduce computation time and the number of generated spikes.
Formally, the prediction $\hat{y}$ of the SNN is defined as:
\begin{equation}
    \begin{aligned}
        \hat{y} & = c_{j^*}                                           \\
        j^*     & = \argmin_{j \in \left[1,N\right]}\left(t_j\right),
    \end{aligned}
\end{equation}
where $t_j$ denotes the firing timestamp of neuron $n_j$.
If several neurons fire at the same timestamp, the one with the highest membrane potential is selected.
In practice, the method used for selecting a neuron in the event of a tie has a negligible impact on the performance.

\subsection{Stabilized Supervised STDP} \label{sec:s2stdp}
To optimize the synaptic weights of the classification layer, we propose an error-modulated supervised STDP learning rule named Stabilized Supervised STDP (S2-STDP).
This rule teaches neurons to alternate their firing between two desired timestamps, $T_\mathrm{target}$ and $T_\mathrm{non-target}$, dynamically computed for each sample based on $T_{\mathrm{mean}}$, the average firing time in the layer.
If the class of the neuron corresponds to the class of the sample, the neuron receives a target weight update, teaching it to fire closer to $T_\mathrm{target}$, just before $T_{\mathrm{mean}}$.
If the class of the neuron does not correspond to the class of the sample, the neuron receives a non-target weight update, teaching it to fire closer to $T_\mathrm{non-target}$, right after $T_{\mathrm{mean}}$.
During training, at the end of the presentation of a sample, the weights of each output neuron are updated with an error-modulated adaptation of the multiplicative STDP:
\begin{equation}
    \Delta w_{ij} =
    \begin{cases}
        e_j \times A^{+} \times \exp\left(-\beta\frac{ w_{ij}-w_{\mathrm{min}}}{w_{\mathrm{max}}-w_{\mathrm{min}}}\right) & \text{if } t_j \geq t_i \\
        e_j \times A^{-} \times \exp\left(-\beta\frac{w_{\mathrm{max}}-w_{ij}}{w_{\mathrm{max}}-w_{\mathrm{min}}}\right)  & \text{o.w. }
    \end{cases},
    \label{eq:s2stdp-update}
\end{equation}
where $w_{ij}$ is the weight of the synapse connecting input neuron $n_i$ and output neuron $n_j$, $\Delta w_{ij}$ is the weight change (such as  $w_{ij} = w_{ij} + \Delta w_{ij}$), $e_j$ is the error of neuron $n_j$, $A^+$ and $A^-$ are the positive and negative learning rates, $\beta$ is the saturation factor, $w_{\mathrm{min}}$ and $w_{\mathrm{max}}$ are the minimum and maximum achievable weight values in the layer, and $t_i$ is the firing timestamp of neuron $n_i$. Multiplicative STDP reduces the effect of weight saturation by adjusting the update according to the current weight value and boundaries~\citep{querliozSimulationMemristorbasedSpiking2011}.
Regardless, weights are manually clipped in $\left[w_{\mathrm{min}}, w_{\mathrm{max}}\right]$ after each update to ensure that they remain within a controlled range.
The error of an output neuron $n_j$ is defined as a temporal difference:
\begin{equation}
    e_j = \frac{t_j - T_j}{T_{\mathrm{max}}},
\end{equation}
where $t_j$ and $T_j$ respectively represent the actual and desired firing timestamps, and $T_{\mathrm{max}}$ is the maximum firing time.
To compute the desired firing timestamps, we introduce a method adapted from SSTDP:
\begin{equation}
    T_j=\begin{cases}
        \begin{aligned}
            T_\mathrm{target}     & = T_{\mathrm{mean}}-\frac{N-1}{N} \cdot g & \text{if } c_j=y      \\
            T_\mathrm{non-target} & = T_{\mathrm{mean}}+\frac{1}{N} \cdot g   & \text{if } c_j \neq y
        \end{aligned}
    \end{cases},
    \label{eq:s2stdp-tdesired}
\end{equation}
where $T_{\mathrm{mean}}$ is the average firing time in the layer, $N$ is the number of neurons, $c_j$ is the class of neuron $n_j$, $y$ is the class of the input sample, and $g$ is a time gap hyperparameter that determines the desired distance from $T_{\mathrm{mean}}$.
Specifically, whereas SSTDP defines desired time ranges $\left[0, T_\mathrm{target}\right]$ and $\left[T_\mathrm{non-target}, T_{\mathrm{max}}\right]$, our adaptation defines desired timestamps $T_\mathrm{target}$ and $T_\mathrm{non-target}$.
Therefore, with our adaptation, neurons can undergo weight updates in both directions, regardless of their associated class.
For instance, if a neuron that does match the input class fires before its desired timestamp ($T_\mathrm{target}$), its weights will receive a negative update to promote later firing.
Conversely, if a neuron that does not match the input class fires after its desired timestamp ($T_\mathrm{non-target}$), its weights will receive a positive update to promote earlier firing.
With SSTDP, the weights of these neurons would not be updated in such cases.

S2-STDP is carefully designed to resolve the two issues of SSTDP outlined in Section~\ref{sec:Problem}.
The first aim of S2-STDP is to increase the number of STDP updates per epoch to facilitate training convergence at higher accuracy.
This is addressed by defining floating-point desired timestamps rather than time ranges, which are considerably more challenging for neurons to reach.
The second aim of S2-STDP is to reduce the saturation of firing timestamps to improve the expressivity of the SNN and its ability to separate classes.
This is addressed by enabling positive non-target weight updates that promote earlier firing, and hence, stabilize the output spikes at earlier timestamps.
Earlier output firing timestamps may allow the SNN to better fit the specificity of input spikes from a certain class, resulting in a higher variance in the average firing time across samples of different classes.
Note that, compared to SSTDP, S2-STDP pushes neurons to fire closer to $T_{\mathrm{mean}}$, which may reduce the variance between neuron firing timestamps for a given sample.
However, we argue that this is not a concern since the desired timestamps provide more accurate control over the output firing timestamps.

In addition to STDP, we use a heterosynaptic plasticity model~\citep{ferreUnsupervisedFeatureLearning2018,liangImpactEncodingDecoding2018} to regulate changes in synaptic weights.
This model ensures that all neurons maintain a constant and similar weight average throughout the training process, allowing them equal chances of activation regardless of the number of weight updates they have undergone.
After each update of an output neuron $n_j$, its weights are normalized as follows:
\begin{equation}
    w_{ij}=w_{ij} \cdot \frac{f_{\mathrm{norm}}}{\sum_{k}^{} w_{kj}},
\end{equation}
where $w_{ij}$ represents the weight of the synapse with input neuron $n_i$, and $f_{\mathrm{norm}}$ is the normalization factor, computed as the sum of weights of neuron $n_j$ at initialization.
In practice, this method also provides robustness against the choice of positive and negative learning rates ($A^{+}$ and $A^{-}$).

\subsection{Paired Competing Neurons}

\begin{figure}[!ht]
    \begin{center}
        \includegraphics[width=\textwidth]{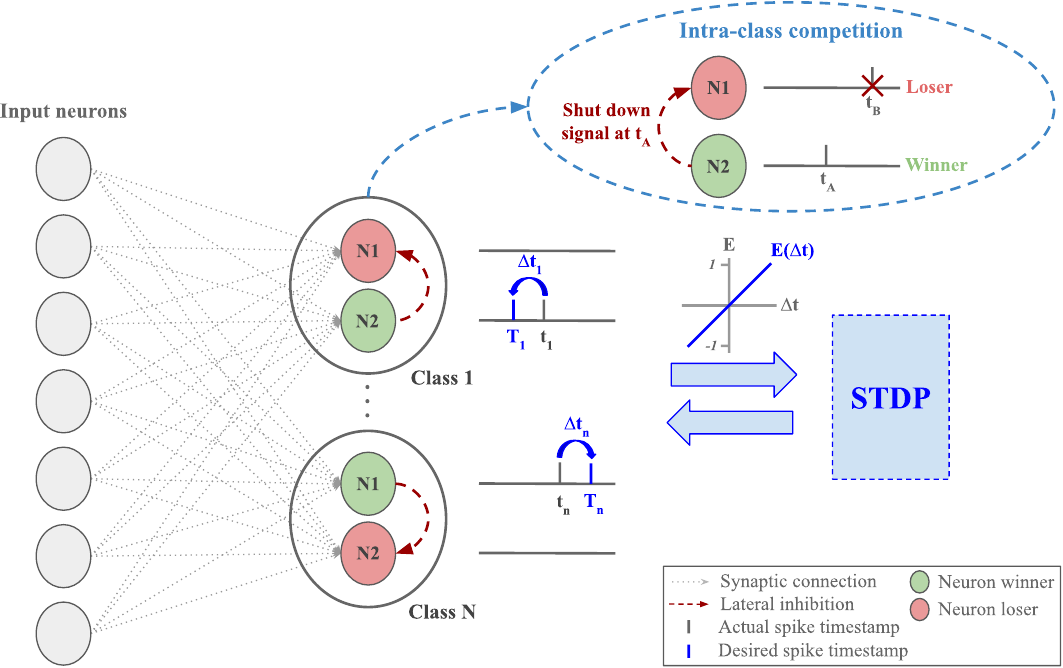}
    \end{center}
    \caption{Classification layer equipped with Paired Competing Neurons (PCN) and trained via Stabilized Supervised STDP (S2-STDP). Each class is represented by paired neurons, interconnected with lateral inhibition to create intra-class competition. Within a pair, the first neuron to fire (the winner) inhibits the other one (the loser) and undergoes the STDP update. The difference $\Delta t$ between the desired and actual firing timestamps is used to compute the neuron error, which modulates the intensity and the polarity of the STDP update. The purpose of PCN is to enhance the learning capabilities of the neurons by promoting specialization toward target or non-target samples.}

    \label{fig:pcn}
\end{figure}
S2-STDP involves training each neuron to alternate its firing between two desired timestamps, depending on the class of the input sample (see Equation~\ref{eq:s2stdp-tdesired}).
Hence, the neurons have strong training requirements: they must adapt their weights to satisfy both target and non-target desired timestamps.
These requirements can limit their learning capabilities (i.e. their capabilities to learn class-specific patterns) because it is harder to find weight values that allow them to reach both desired timestamps.
However, ensuring the convergence toward the desired timestamps is of crucial importance to stabilize the output firing timestamps in the layer.
To benefit from the stabilized property of S2-STDP along with enhanced learning capabilities, we propose the Paired Competing Neurons (PCN) training architecture, described in Figure~\ref{fig:pcn}.

In this architecture, each class $c_{j \in \left[1, N\right]}$ is associated with a pair of output neurons $\left(n_j, n_j\prime\right)$ that are connected with lateral inhibition to create intra-class competition.
Within each pair and for each sample, the first neuron to fire, called the winner, inhibits the other one, called the loser, and undergoes the weight update.
This mechanism is similar to the WTA competition found in STDP-based networks, yet it is class-specific in this context.
The purpose of PCN is to encourage, for each class, neuron specialization toward one type of sample: target or non-target.
In other words, for a pair of neurons $\left(n_j, n_j\prime\right)$ associated with class $c_j$, when we present samples of class $y = c_j$, we want $n_j$ to fire at $T_\mathrm{target}$ and $n_j\prime$ to fire after $n_j$.
Conversely, when we present samples of class $y \neq c_j$, we want $n_j\prime$ to fire at $T_\mathrm{non-target}$ and $n_j$ to fire after $n_j\prime$.
The neuron order in this example is arbitrary as we do not assign an objective to each neuron, i.e., within a class-specific pair, we do not explicitly label neurons as target or non-target.
Instead, we let this behavior emerge naturally through intra-class competition.
Note that the purpose of neurons specializing toward non-target samples extends beyond inhibiting target firing for non-target samples~\citep{tavanaeiMinimalSpikingNeural2015}.
Here, these neurons also play a crucial role in the S2-STDP training process, as their firing timestamps directly impact the average firing time, and hence, the training convergence.
By encouraging specialization toward one type of sample, we reduce training requirements on the neurons because their weights primarily receive one type of update, which improves their learning capabilities.

It should be noted that the use of PCN offers notable other advantages.
First, thanks to class-wise lateral inhibition, the increased number of neurons does not introduce additional training complexity because, for each sample, only the winners receive weight updates (i.e. one neuron per class).
Second, it does not introduce any additional hyperparameters.
Third, it can, in principle, be used with any other learning rules involving two desired timestamps.

\section{Results} \label{sec:Results}

\subsection{Experimental setup}

\subsubsection{Feature extraction network}
Before classification, we extract features from images with a CSNN trained using unsupervised STDP.
In our experiments, we use the CSNN model of~\cite{falezMultilayeredSpikingNeural2019}, described in Section~\ref{sec:CSNN}.
The employed architecture consists of a single trainable convolutional layer followed by a non-trainable max-pooling layer.
Training is done in a layer-wise fashion: the convolutional layer of the CSNN is entirely trained before the training of the classification layer starts.
Additional training details are provided in Supplementary Material (Section~1).
To analyze the performance of the classification layer across various input sizes, we consider three configurations of the CSNN with increasing numbers of filters: CSNN-16 (16 filters), CSNN-64 (64 filters), and CSNN-128 (128 filters).
Within a given dataset, these configurations share the same hyperparameters, except for the number of filters.
Unless otherwise specified, the experiments are conducted on CSNN-128.

\subsubsection{Datasets}
We select three image recognition datasets, each comprising ten classes, exhibiting growing complexity: MNIST~\citep{lecunGradientbasedLearningApplied1998}, Fashion-MNIST~\citep{xiaoFashionMNISTNovelImage2017}, and CIFAR-10~\citep{krizhevskyLearningMultipleLayers2009}.
Both MNIST and Fashion-MNIST consist of $28\times28$ grayscale images.
They contain 60,000 samples for training and 10,000 samples for testing.
We preprocess the images with on-center/off-center coding to extract edge information~\citep{falezMultilayeredSpikingNeural2019}.
CIFAR-10 is composed of $32\times32$ RGB images, 50,000 for training and 10,000 for testing.
We preprocess the images with the hardware-friendly whitening method presented in~\citep{falezImprovingSTDPbasedVisual2020} to highlight their edges and high-frequency features.
Note that CIFAR-10 is challenging for STDP-based SNNs and only a limited number of studies have considered this dataset thus far~\citep{ferreUnsupervisedFeatureLearning2018,srinivasanReStoCNetResidualStochastic2019,falezImprovingSTDPbasedVisual2020,shresthaInHardwareLearningMultilayer2021}.
All the preprocessing methods are used with their original hyperparameters, provided in Supplementary Material (Section~2).

\subsubsection{Protocol}
We divide our experimental protocol into two phases: hyperparameter optimization and evaluation.
In both phases, we employ an early stopping mechanism (with a patience $p_{\mathrm{stop}}$) during training to prevent overfitting.

During the hyperparameter optimization phase, a subset of the training set is used for validation, which is created by randomly selecting, for each class, a percentage $p_{\mathrm{val}}$ of its samples.
We then apply the gridsearch algorithm to optimize the hyperparameters of the SNN based on the validation accuracy.
Hence, we ensure that the hyperparameters are not optimized for the test set.
In this work, only the hyperparameters of the classification layer are optimized with gridsearch.

The hyperparameters of the CSNN are manually set based on preliminary experiments.
For each dataset and model of the classification layer, gridsearch is performed on the CSNN-128 configuration.
Then, the same hyperparameters are used with CSNN-64 and CSNN-16, except for the firing threshold that is divided by 2 and 4, respectively, as the number of input spikes decreases with the input size.
All the hyperparameters are provided in Supplementary Material (Section~2).

During the evaluation phase, we use the K-fold cross-validation strategy.
We partition the training set into $K$ subsets and train $K$ models, each using a different subset for validation while the remaining $K-1$ subsets are used for training.
Then, we evaluate the trained models on the test set and we compute the average test accuracy.
Note that each model is trained with a different random seed.
This allows us to assess the performance of the SNN with varying training and validation data, as well as different weight initializations.
In all the following experiments, we choose $p_{\mathrm{stop}}=10$, $K=10$ and $p_{\mathrm{val}}=\frac{1}{K}$ (i.e. $10$\% of the training set is used for validation during the hyperparameter optimization phase).

\subsection{Accuracy comparison} \label{sec:results-acc-comparison}
In this section, we present a comparative analysis of the accuracy between two existing methods, R-STDP and SSTDP (both of which we have implemented and optimized using gridsearch), along with our methods, S2-STDP and S2-STDP+PCN.
Tables~\ref{tab:acc-csnn-mnist},~\ref{tab:acc-csnn-fmnist}, and~\ref{tab:acc-csnn-cifar10} show the average test accuracy achieved by each method on the MNIST, Fashion-MNIST, and CIFAR-10 datasets, respectively.

\begin{table}[!ht]
    \def\arraystretch{1.5}
    \caption{Accuracy of the existing and proposed learning rules on MNIST.}
    \centering
    \begin{tabular}{|c|c|c|c|c|}
        \hline
        \multirow{2}{*}{Rule} & \multirow{2}{*}{\makecell{Output                                                                                                                \\ Neurons}} &  \multirow{2}{*}{CSNN-16}                            & \multirow{2}{*}{CSNN-64}                            & \multirow{2}{*}{CSNN-128}                           \\
                              &                                  &                                    &                                    &                                    \\
        \hline
        R-STDP                & 200                              & $96.05 \pm 0.51$                   & $97.47 \pm 0.16$                   & $97.88 \pm 0.13$                   \\
        \hline
        R-STDP                & 20                               & $61.23 \pm 27.74$                  & $93.34 \pm 0.66$                   & $93.28 \pm 0.87$                   \\
        \hline
        SSTDP                 & 10                               & $84.49 \pm 7.07$                   & $95.56 \pm 0.13$                   & $96.25 \pm 0.11$                   \\
        \hline
        S2-STDP*              & 10                               & $95.40 \pm 0.21$                   & $97.19 \pm 0.09$                   & $97.81 \pm 0.05$                   \\
        \hline
        \textbf{S2-STDP+PCN}* & \textbf{20}                      & \textbf{97.02} $\pm$ \textbf{0.14} & \textbf{98.21} $\pm$ \textbf{0.07} & \textbf{98.59} $\pm$ \textbf{0.06} \\
        \hline
    \end{tabular}
    \label{tab:acc-csnn-mnist}
    \\
    \small{* Ours}
\end{table}

\begin{table}[!ht]
    \def\arraystretch{1.5}
    \caption{Accuracy of the existing and proposed learning rules on Fashion-MNIST.}
    \centering
    \begin{tabular}{|c|c|c|c|c|}
        \hline
        \multirow{2}{*}{Rule} & \multirow{2}{*}{\makecell{Output                                                                                                                \\ Neurons}} &  \multirow{2}{*}{CSNN-16}                            & \multirow{2}{*}{CSNN-64}                            & \multirow{2}{*}{CSNN-128}                           \\
                              &                                  &                                    &                                    &                                    \\
        \hline
        R-STDP                & 200                              & $82.30 \pm 0.26$                   & $82.30 \pm 0.92$                   & $83.26 \pm 0.22$                   \\
        \hline
        R-STDP                & 20                               & $76.50 \pm 0.52$                   & $76.73 \pm 0.13$                   & $77.01 \pm 0.22$                   \\
        \hline
        SSTDP                 & 10                               & $81.40 \pm 0.67$                   & $84.51 \pm 0.18$                   & $85.16 \pm 0.10$                   \\
        \hline
        S2-STDP*              & 10                               & $82.23 \pm 0.40$                   & $84.92 \pm 0.24$                   & $85.88 \pm 0.22$                   \\
        \hline
        \textbf{S2-STDP+PCN}* & \textbf{20}                      & \textbf{83.89} $\pm$ \textbf{0.40} & \textbf{85.84} $\pm$ \textbf{0.19} & \textbf{87.12} $\pm$ \textbf{0.21} \\
        \hline
    \end{tabular}
    \label{tab:acc-csnn-fmnist}
    \\
    \small{* Ours}
\end{table}

\begin{table}[!ht]
    \def\arraystretch{1.5}
    \caption{Accuracy of the existing and proposed learning rules on CIFAR-10.}
    \centering
    \begin{tabular}{|c|c|c|c|c|}
        \hline
        \multirow{2}{*}{Rule} & \multirow{2}{*}{\makecell{Output                                                                                                                \\ Neurons}} &  \multirow{2}{*}{CSNN-16}                            & \multirow{2}{*}{CSNN-64}                            & \multirow{2}{*}{CSNN-128}                           \\
                              &                                  &                                    &                                    &                                    \\
        \hline
        \textbf{R-STDP}       & \textbf{200}                     & \textbf{51.55} $\pm$ \textbf{1.23} & \textbf{61.85} $\pm$ \textbf{0.61} & \textbf{65.56} $\pm$ \textbf{0.38} \\
        \hline
        R-STDP                & 20                               & $38.51 \pm 4.41$                   & $51.74 \pm 0.93$                   & $54.02 \pm 0.80$                   \\
        \hline
        SSTDP                 & 10                               & $37.78 \pm 2.41$                   & $57.53 \pm 0.38$                   & $60.88 \pm 0.23$                   \\
        \hline
        S2-STDP*              & 10                               & $47.94 \pm 0.49$                   & $58.24 \pm 0.27$                   & $61.53 \pm 0.16$                   \\
        \hline
        S2-STDP+PCN*          & 20                               & $49.23 \pm 0.56$                   & $59.58 \pm 0.16$                   & $62.81 \pm 0.15$                   \\
        \hline
    \end{tabular}
    \label{tab:acc-csnn-cifar10}
    \\
    \small{* Ours}
\end{table}


Across all datasets and CSNN configurations, S2-STDP consistently outperforms SSTDP.
On CSNN-128, we measure a gain of $1.56$~pp for MNIST, $0.72$~pp for Fashion-MNIST, and $0.65$~pp for CIFAR-10.
In addition, SSTDP tends to underperform when confronted with smaller input sizes.
For instance, on CSNN-16, S2-STDP outperforms SSTDP by $10.91$~pp on MNIST and $10.16$~pp on CIFAR-10.
The lower performance of SSTDP on the CSNN-16 configuration of these datasets is caused by training divergence, as evidenced by the higher standard deviations, and is influenced by the employed hyperparameters (transferred from CSNN-128).
It suggests that SSTDP lacks robustness against hyperparameters.
While the accuracy gain between SSTDP and S2-STDP is not always substantial, our adaptation enables more effective training of the classification layer, irrespective of the number of input features.
More importantly, S2-STDP leverages compatibility with the PCN architecture, which further improves the accuracy of S2-STDP across all datasets and CSNN configurations, without requiring any additional hyperparameters.
When integrating PCN with S2-STDP on CSNN-128, we measure an additional gain of $0.78$~pp for MNIST, $1.24$~pp for Fashion-MNIST, and $1.28$~pp for CIFAR-10.
In comparison to the other existing STDP-based methods, S2-STDP+PCN achieves the highest accuracy on the MNIST and Fashion-MNIST datasets.
Specifically, compared to SSTDP on CSNN-128, it shows an accuracy improvement of $2.34$~pp on MNIST, $1.96$~pp on Fashion-MNIST, and $1.93$~pp on CIFAR-10.
We recall that our proposed training methods based on S2-STDP employ a weight normalization mechanism not employed by SSTDP.
We evaluate the effect of weight normalization in Supplementary Material (Section~3.3).

On CIFAR-10, R-STDP significantly outperforms all the STDP-based methods.
However, R-STDP requires 200 output neurons to achieve this performance whereas S2-STDP+PCN only uses 20 neurons.
When R-STDP is used with 20 output neurons as S2-STDP+PCN, it performs significantly worse than all other methods across all datasets and CSNN configurations.
This observation highlights the importance of error-modulated weight updates in enabling effective and efficient supervised training with STDP.
In our case, it leads to a substantial reduction in the number of trainable parameters by a factor of 10.
Also, on Fashion-MNIST, R-STDP obtains relatively low performance and fails to extract more relevant features when the number of feature maps increases.

The accuracy gain between CSNN-16 and CSNN-128 with R-STDP is only about $0.96$~pp, whereas it is about $3.23$~pp with S2-STDP+PCN.

\subsection{S2-STDP addresses the issues of SSTDP}
In Section~\ref{sec:Problem}, we elaborated on the issues of SSTDP: the limited number of STDP updates per epoch and the saturation of firing timestamps toward the maximum firing time.
Our proposed S2-STDP is specifically designed to address these issues by defining desired timestamps that stabilize the output spikes at earlier timestamps.
In this section, we analyze the effect of S2-STDP on these issues.

\begin{figure}[!ht]
    \centering
    \includegraphics[width=0.8\textwidth]{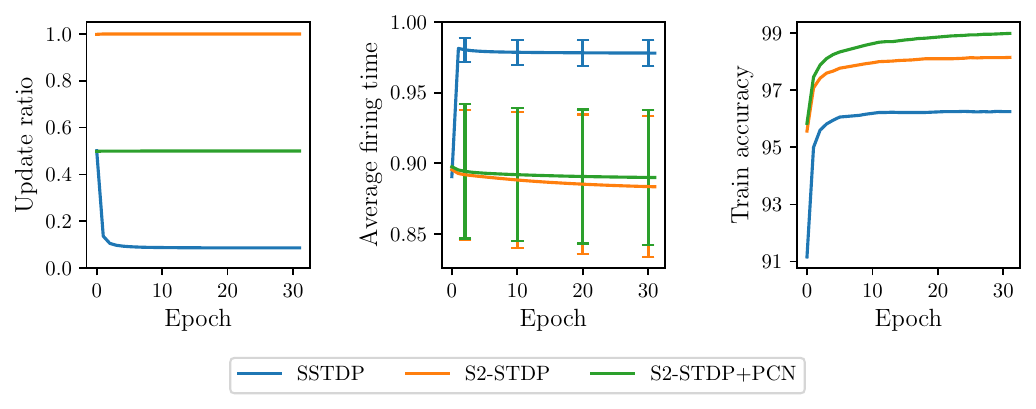}
    \caption{Update ratio, average firing time, and train accuracy per epoch in the classification layer trained on MNIST. Our methods using S2-STDP significantly increase the number of updates per epoch and reduce the saturation of firing timestamps toward the maximum firing time. As a result, they enable training convergence at higher accuracies compared to SSTDP.}
    \label{fig:sstdp_vs_s2stdp_mnist}
\end{figure}
In Figure~\ref{fig:sstdp_vs_s2stdp_mnist}, we compare the update ratio, average firing time, and train accuracy per epoch in the classification layer trained with the various SSTDP-based methods on the MNIST dataset.
As previously illustrated in a preliminary experiment (see Figures~\ref{fig:sstdp_nb_updates} and~\ref{fig:sstdp_tmean_evol}), SSTDP suffers from a limited number of updates per epoch, firing timestamps close to the maximum firing time, and low variance in the average firing time between samples.
Our proposed methods based on S2-STDP successfully address these three issues.
In comparison to SSTDP at epoch 30, S2-STDP increases the update ratio from $9$\% to nearly $100$\%, reduces the average firing time from $0.98$ to $0.88$, and augments its standard deviation from $0.009$ to $0.05$.
Note that with S2-STDP+PCN, the update ratio is close to $50$\% (instead of $100$\%), which is the maximum achievable value because half of the neurons are inhibited.
By addressing the issues of SSTDP, our methods based on S2-STDP enable training convergence at higher accuracies.
In addition, due to the increased number of updates per epoch, they achieve higher accuracies in fewer epochs.
For instance, S2-STDP reaches a training accuracy of $95.56$\% after the first epoch, whereas SSTDP only reaches $91.16$\% and requires two additional epochs to reach $95.60$\%.
In Supplementary Material (Section~3.1), we show similar results for the Fashion-MNIST dataset and we demonstrate that the resolution of these issues is not attributed to the additional weight normalization mechanism employed.

\begin{figure}[!ht]
    \centering

    \includegraphics[width=0.8\textwidth]{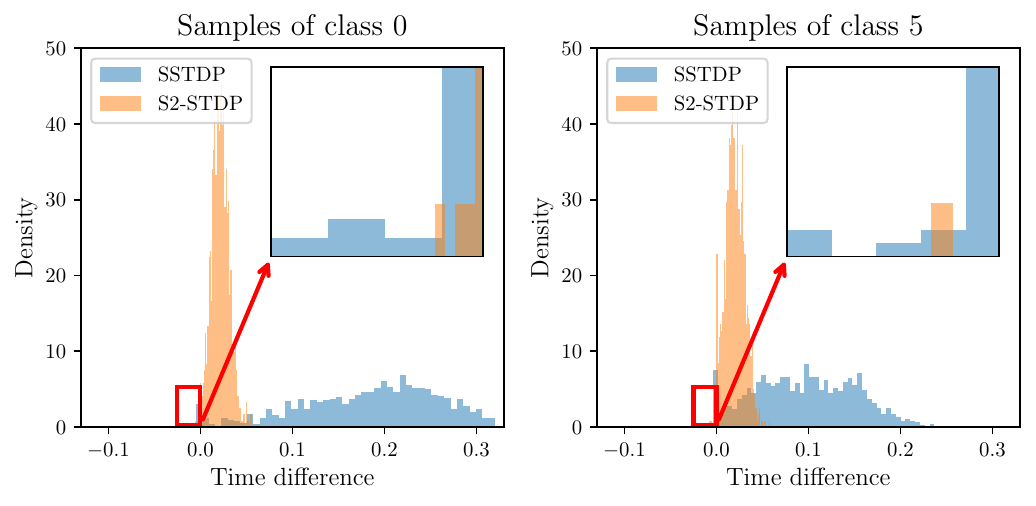}
    \caption{Distribution of firing time differences between the first non-target neuron to fire and the target neuron, in the classification layer for MNIST test samples of class 0 and 5. A negative time difference indicates that the target neuron fired after a non-target neuron, leading to misclassification of the sample. S2-STDP achieves a tighter distribution of firing time differences but results in fewer misclassified samples compared to SSTDP.}
    \label{fig:mnist_spk_diff_t_nt}

\end{figure}
We mentioned in Section~\ref{sec:s2stdp} that S2-STDP pushes neurons to fire closer to the average firing time, and hence, to each other, compared to SSTDP.
In the following experiment, we show that contrary to intuition, a tighter distribution of output firing timestamps does not necessarily reduce the ability of the SNN to separate classes.
Figure~\ref{fig:mnist_spk_diff_t_nt} illustrates the distribution of firing time differences between the first non-target neuron to fire and the target neuron, on MNIST test samples.
Since the SNN prediction is based on the neuron that fires first, the sign of the time difference indicates the ability of the SNN to classify the sample: a positive time difference indicates a correctly classified sample (the target neuron fires before the non-target neuron), and conversely.
The tight distribution of S2-STDP implies that the firing time differences tend to be significantly smaller compared to SSTDP.
This behavior arises from our adapted training process, which pushes neurons to fire closer to the average firing time.
While it may seem intuitive to maximize the firing time difference for improved class separability, a closer examination of negative time differences reveals that SSTDP leads to more misclassified samples than S2-STDP (which is confirmed by the overall lower accuracy of SSTDP).
In the context of temporal decision-making, we argue that supervised STDP rules should not necessarily aim to maximize the firing time difference between the target and non-target neurons.
Instead, it seems more important to ensure accurate control over the output firing timestamps.
In Supplementary Material (Section~3.1), we present similar results with the other classes of the MNIST dataset.

\subsection{PCN enables neuron specialization}
The PCN architecture exploits the two desired timestamps defined by S2-STDP along with intra-class competition to promote neuron specialization toward one type of sample: target or non-target.
In this section, we study the impact of integrating a PCN architecture on the output firing timestamps of our classification layer trained with S2-STDP.

\begin{figure}[!ht]
    \centering

    \includegraphics[width=\textwidth]{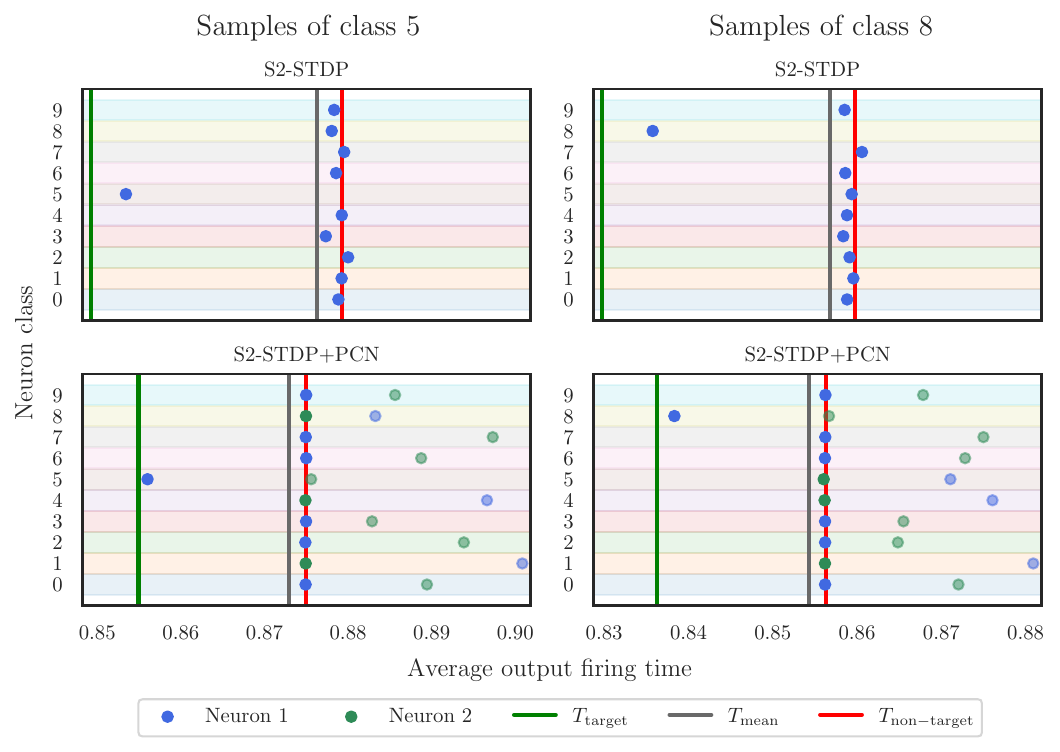}
    \caption{Neuron average firing time in the classification layer, for MNIST test samples of class 5 and 8. Each point represents a neuron, and the row denotes its associated class. The points with higher transparency are the inhibited neurons (i.e. losers). The green and red lines are the average desired timestamps of the target neuron and the non-target neurons, respectively. Through intra-class competition, the use of the PCN architecture enables neuron specialization toward one type of sample, which helps them reach their desired firing timestamp and improve their learning capabilities.}

    \label{fig:mnist_out_spks}
\end{figure}
Figure~\ref{fig:mnist_out_spks} illustrates the average firing time of output neurons trained with and without PCN, on MNIST test samples of classes 5 and 8.
Note that the time gaps used for S2-STDP and S2-STDP+PCN differ but they correspond to the optimal value obtained through gridsearch.
We observe that neurons trained with a PCN architecture better reach their desired timestamps, particularly for the non-target neurons (see Section~3.2 of Supplementary Material for additional analysis).
This is because, through intra-class competition, PCN naturally enables neuron specialization toward target or non-target samples.
This specialization is illustrated by neuron 1 of class 5 firing at $T_\mathrm{target}$ when presented to samples of class 5 (target class) and being inhibited by neuron 2 of class 5 for samples of class 8 (non-target class).
Similarly, neuron 2 of class 8 fires at $T_\mathrm{non-target}$ when presented to samples of class 5 (non-target class) and is inhibited by neuron 1 of class 8 for samples of class 8 (target class).
Intra-class competition reduces training requirements on neurons specialized toward target samples because they do not have to fire at an exact desired timestamp ($T_\mathrm{non-target}$) for non-target samples.
As observed, inhibited neurons for non-target samples (i.e. neurons specialized toward target samples) have much larger firing timestamps compared to the non-target desired timestamp.
Consequently, by creating neuron specialization toward target or non-target samples, PCN facilitates the learning of more specific patterns, which improves their learning capabilities.

\subsection{Robustness against the time gap hyperparameter}
During training, the time gap hyperparameter $g$ of S2-STDP is used to define the distance between the desired timestamps and the average firing time.
Selecting an appropriate value for this hyperparameter is crucial to ensure accurate class separation.
However, hyperparameter tuning can be a time-consuming task, and achieving the optimal value may not always be feasible.
In this section, we investigate the influence of the time gap value on accuracy.
\begin{figure}[!ht]
    \centering

    \includegraphics[width=0.8\textwidth]{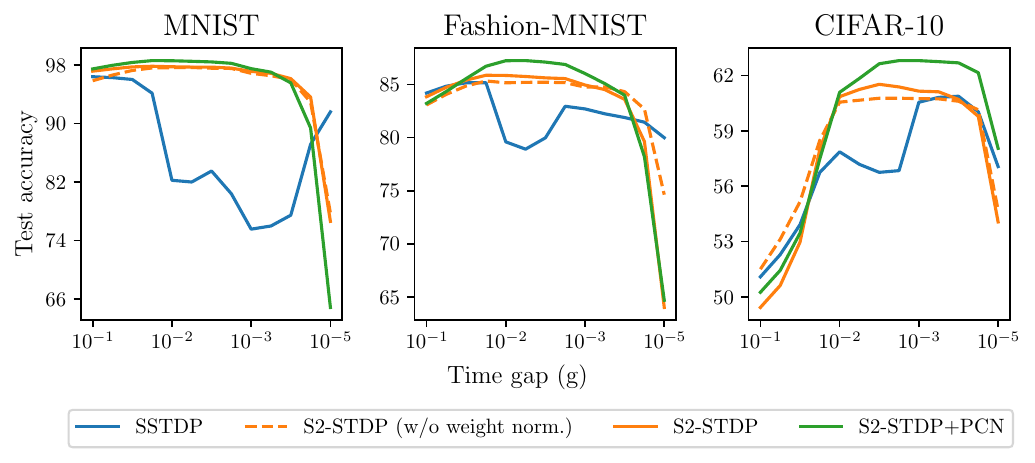}
    \caption{Accuracy of the different SSTDP-based learning rules against the time gap hyperparameter. Our proposed methods using S2-STDP enable better robustness against the time gap value.}
    \label{fig:t_gap_analysis}
\end{figure}

Figure~\ref{fig:t_gap_analysis} compares SSTDP and our proposed methods across various values of $g$ on the three datasets.
S2-STDP demonstrates greater robustness compared to SSTDP regarding the choice of $g$, as it significantly expands the range of values that can achieve near-optimal accuracy.
The accuracy curve of S2-STDP exhibits a more pronounced bell-shaped pattern with a larger plateau near the maximum.
This implies that tuning $g$ can be easier with S2-STDP.
When considering a suitable range for $g$, S2-STDP always achieves higher or similar accuracy compared to SSTDP.
We also illustrate the accuracy of S2-STDP without weight normalization to demonstrate that its improved robustness is not due to this additional mechanism.
The use of PCN as a training architecture for S2-STDP almost always improves its performance.
Hence, the efficacy of PCN is not dependent on a specific value of $g$.
Note that all the methods use their respective gridsearch-optimized hyperparameters.

\subsection{Robustness against the hyperparameter set}
Our PCN training architecture improves the accuracy of S2-STDP across all evaluated datasets, as indicated in Section~\ref{sec:results-acc-comparison}.
However, the hyperparameters of the classification layer are individually optimized for S2-STDP and S2-STDP+PCN, suggesting that PCN might only enhance performance with specific hyperparameter sets.
In this section, we present evidence that PCN can effectively improve the accuracy of S2-STDP irrespective of the hyperparameters used.
Specifically, we aim to demonstrate that integrating PCN into an existing SNN employing S2-STDP for training is likely to always yield improved results.
\begin{figure}[!ht]
    \centering

    \includegraphics[width=\textwidth]{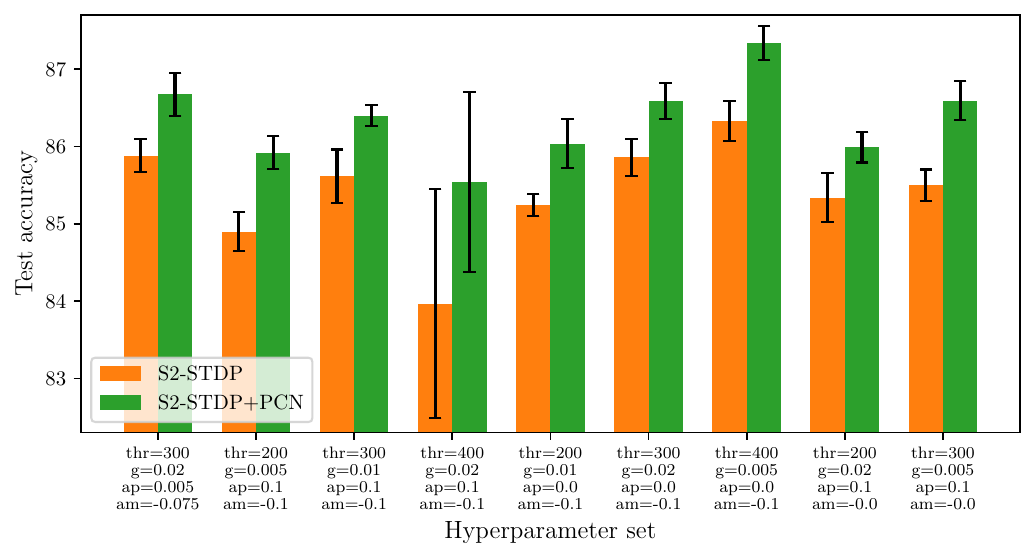}
    \caption{Accuracy of S2-STDP, with and without the use of the PCN architecture, across different hyperparameter sets on Fashion-MNIST. PCN always improves S2-STDP performance, without introducing any additional hyperparameters.}
    \label{fig:fmnist-config_analysis}
\end{figure}

Figure~\ref{fig:fmnist-config_analysis} illustrates the comparison of S2-STDP accuracy with and without PCN across different hyperparameter sets on the Fashion-MNIST dataset.
Results on the other datasets are reported in Supplementary Material (Section~3.4).
The varying hyperparameters include the firing threshold ($\mathrm{thr}$), time gap ($g$), and learning rates ($\mathrm{ap}$ and $\mathrm{am}$).

Their values are selected within a suitable range to achieve satisfactory performance with S2-STDP.
The results show that integrating PCN improves the accuracy of S2-STDP consistently, with an average improvement of $0.94$~pp and a maximum improvement of $1.57$~pp.
For comparison, when independently optimizing the hyperparameters for both methods, the measured accuracy improvement is $1.24$~pp.
Hence, the PCN architecture is an effective method for easily improving the performance of a spiking classifier trained with S2-STDP, regardless of the hyperparameter set and without introducing any additional hyperparameters.
As detailed in Supplementary Material, the results on the other datasets exhibit consistency with the analysis conducted on Fashion-MNIST.

\subsection{Comparison with the literature}
In Table~\ref{tab:sota-acc}, we present an accuracy comparison between our partially supervised SNN, trained with STDP and S2-STDP+PCN, along with other existing algorithms employed for training SNNs.
For this comparison, we focused on supervised methods, primarily STDP-based and BP-based with local updates.
Note that the reported accuracies come from the original papers.

\begin{table}[!ht]
    \footnotesize
    \caption{Accuracy comparison of our proposed SNN with the literature. Accuracies come from the original papers.}
    \hspace*{-0.5cm}
    \begin{tabularx}{\linewidth}{ccccCC}

        \hline
        \multirow{2}{*}{Dataset} & \multirow{2}{*}{Model}                                       & \multirow{2}{*}{\makecell{Number of layers                                                                  \\ Unsupervised / Supervised}} & \multirow{2}{*}{Learning rule}                  & \multirow{2}{*}{Local}        & \multirow{2}{*}{Accuracy}       \\
                                 &                                                              &                                            &                                &              &                \\
        \hline
        MNIST                    & \citep{leeTrainingDeepSpiking2018}                           & \textbf{0 / 4}                             & \textbf{STDP pretraining + BP} & \textbf{no}  & \textbf{99.28} \\
                                 & \citep{liuSSTDPSupervisedSpike2021}                          & 0 / 2                                      & SSTDP                          & no           & $98.10$        \\
                                 & \citep{tavanaeiBPSTDPApproximatingBackpropagation2019}       & 0 / 3                                      & BP-STDP                        & no           & $97.20$        \\
                                 & \citep{zhangTuningConvolutionalSpiking2021}                  & 0 / 3                                      & BRP                            & yes          & $99.01$        \\
                                 & \citep{shresthaApproximatingBackpropagationBiologically2019} & 0 / 3                                      & EMSTDP                         & yes          & $97.30$        \\
                                 & \citep{mozafariBioinspiredDigitRecognition2019}              & 2 / 1                                      & STDP + R-STDP                  & yes          & $97.20$        \\
                                 & \textit{This work}                                           & \textit{1} / \textit{1}                    & \textit{STDP + S2-STDP+PCN}    & yes          & \textit{98.59} \\
        \hline
        Fashion-MNIST            & \citep{kheradpishehTemporalBackpropagationSpiking2020}       & 0 / 2                                      & BP                             & no           & $88.00$        \\
                                 & \citep{mirsadeghiSpikeTimeDisplacementbased2023}             & \textbf{0 / 4}                             & \textbf{STiDi-BP}              & \textbf{yes} & \textbf{92.80} \\

                                 & \citep{zhaoGLSNNMultiLayerSpiking2020}                       & 0 / 6                                      & Global feedback + STDP         & yes          & $89.05$        \\
                                 & \citep{shresthaApproximatingBackpropagationBiologically2019} & 0 / 3                                      & EMSTDP                         & yes          & $86.10$        \\
                                 & \citep{haoBiologicallyPlausibleSupervised2020}               & 1 / 1                                      & Sym-STDP                       & yes          & $85.31$        \\
                                 & \textit{This work}                                           & \textit{1} / \textit{1}                    & \textit{STDP + S2-STDP+PCN}    & yes          & \textit{87.12} \\
        \hline
        CIFAR-10                 & \citep{liuSSTDPSupervisedSpike2021}                          & \textbf{0 / 7}                             & \textbf{SSTDP}                 & \textbf{no}  & \textbf{91.31} \\
                                 & \citep{ferreUnsupervisedFeatureLearning2018}                 & 1 / 3                                      & STDP + ANN-based BP            & no           & $71.20$        \\
                                 & \citep{srinivasanReStoCNetResidualStochastic2019}            & 1 / 1                                      & STDP + ANN-based BP            & no           & $66.23$        \\
                                 & \citep{shresthaInHardwareLearningMultilayer2021}             & 0 / 4                                      & EMSTDP                         & yes          & $64.40$        \\
                                 & \citep{zhangTuningConvolutionalSpiking2021}                  & 0 / 3                                      & BRP                            & yes          & $57.08$        \\
                                 & \textit{This work}                                           & \textit{1} / \textit{1}                    & \textit{STDP + S2-STDP+PCN}    & yes          & \textit{62.81} \\
        \hline
    \end{tabularx}
    \label{tab:sota-acc}
\end{table}

On the MNIST dataset, our proposed SNN outperforms the STDP-based approaches and achieves results that are close to the non-local BP-based state of the art.
Similarly, on the Fashion-MNIST dataset, our SNN surpasses the STDP-based approaches with shallow architectures~\citep{haoBiologicallyPlausibleSupervised2020,shresthaApproximatingBackpropagationBiologically2019} and demonstrates competitive performance compared to most of the other methods.
It is important to mention that only the output layer of our SNN is trained with supervision, whereas all the other methods in Fashion-MNIST are fully supervised.
For instance, in~\citep{zhaoGLSNNMultiLayerSpiking2020}, they employ a 6-layer SNN trained with global feedback + STDP, achieving an accuracy of $89.05$\%.
In contrast, we employ a 2-layer partially supervised CSNN, resulting in an accuracy loss of only $1.93$~pp.
Additionally, our results are highly competitive with a 2-layer SNN trained using a BP-based algorithm, with an accuracy loss of $0.88$~pp.

On the CIFAR-10 dataset, the performance of our method remains significantly low compared to state-of-the-art algorithms.
However, these algorithms, like the VGG-7 trained with SSTDP~\citep{liuSSTDPSupervisedSpike2021} (which reports a top-1 accuracy and not an average), rely on non-local learning rules that cannot be employed for direct training on neuromorphic hardware.
On the contrary, our proposed SNN is exclusively trained using local learning rules and restricts the use of supervision to one layer.
In comparison to other local-based methods, our SNN demonstrates competitive results.
Overall, this dataset is particularly challenging for shallow architectures with only one supervised layer.
We believe that the features extracted by our unsupervised layer are not distinguishable enough to allow for an accurate analysis.

\section{Conclusion} \label{sec:Conclusion}
In this paper, we proposed Stabilized Supervised STDP (S2-STDP), a supervised STDP learning rule for training a spiking classification layer with one spike per neuron and temporal decision-making.
This layer can be employed to classify features extracted by a convolutional SNN (CSNN) equipped with unsupervised STDP.
Our learning rule integrates error-modulated weight updates that align neuron spikes with desired timestamps derived from the average firing time within the layer.
Then, to further enhance the learning capabilities of the classification layer trained with S2-STDP, we introduced a training architecture called Paired Competing Neurons (PCN).
PCN associates each class with paired neurons connected via lateral inhibition and encourages neuron specialization through intra-class competition.
We evaluated S2-STDP and PCN on three image recognition datasets of growing complexity: MNIST, Fashion-MNIST, and CIFAR-10.
Experiments showed that our methods outperform state-of-the-art supervised STDP rules when employed to train our spiking classification layer.
S2-STDP successfully addresses the issues of SSTDP concerning the limited number of STDP updates per epoch and the saturation of firing timestamps toward the maximum firing time.
The PCN architecture enhances the performance of S2-STDP, regardless of the hyperparameter set and without introducing any additional hyperparameters.
Our methods also exhibited improved hyperparameter robustness as compared to SSTDP.

In the future, we plan to expand S2-STDP to multi-layer architectures, while maintaining the local computation required for on-chip learning.
This includes exploring both feedback connexions~\citep{zhaoGLSNNMultiLayerSpiking2020} and local losses~\citep{mirsadeghiSTiDiBPSpikeTime2021}.

\section*{Author contributions}
GG designed S2-STDP and the PCN architecture.
GG, PT, and IMB designed the experiments.
GG implemented the models and performed the experiments.
GG wrote the first draft.
GG, PT, and IMB corrected the manuscript.
PT and IMB supervised the study.
IMB provided the funding.

\section*{Funding}
This work is funded by Chaire Luxant-ANVI (Métropole Européenne de Lille) and supported by IRCICA (CNRS UAR 3380).

\section*{Conflict of interest}
The authors declare that the research was conducted in the absence of any commercial or financial relationships that could be construed as a potential conflict of interest.

\section*{Acknowledgements}
We would like to thank Benoit Miramond for the helpful exchange.



\section*{Supplementary material}

# SuppM-I    Training of the feature extraction network

To improve image representation before classification, we use a CSNN trained with unsupervised STDP as a feature extraction network. The trainable component of our CSNN is the spiking convolutional layer. Similar to traditional convolutional neural networks, the neurons in the spiking convolutional layer are arranged in two-dimensional maps. Within each map, neurons share weights and thresholds. To avoid the non-local communication required by shared weights, we employ a specific training procedure, with only one active neuron per map (Falez et al., 2019). For each sample, we first extract a random patch from the input image, matching the dimensions of the kernel. Then, we propagate the encoded patch to the active neurons of the convolutional layer. Due to the WTA competition, only the first neuron to fire is updated with the multiplicative STDP (described in the main paper). Then, the thresholds of the active neurons are updated through two adaptation rules. The first one is applied to control the timestamp at which they should fire, which impacts the learnable pattern granularity:

$$V_{\text{th}_i} = \max \left\{ \text{Th}_{\min}, V_{\text{th}_i} - \eta_{\text{th}} \left( t_i - t_{\text{target}} \right) \right\},$$

where $V_{\text{th}_i}$ represents the threshold of neuron $n_i$, $\text{Th}_{\min}$ is the minimum threshold, $\eta_{\text{th}}$ is the learning rate, $t_i$ and $t_{\text{target}}$ are the actual and target firing timestamp of the neuron. The second one is applied to ensure homeostasis (i.e. fair competition among neurons) in the layer:

$$\Delta V_{\text{th}_i} = \begin{cases} \eta_{\text{th}} & \text{if } t_i = \min \left\{ t_1, \cdots, t_N \right\} \\ \frac{-\eta_{\text{th}}}{N} & \text{o.w.} \end{cases}$$

$$V_{\text{th}_i} = \max \left\{ \text{Th}_{\min}, V_{\text{th}_i} + \Delta V_{\text{th}_i} \right\},$$

where $N$ is the number of neurons in competition (i.e. the number of maps). Once the layer is trained, the parameters (weights and thresholds) of the active neurons are fixed and are copied onto the other neurons of the maps.

# SuppM-II    hyperparameters

To preprocess MNIST and Fashion-MNIST datasets, we use on-center/off-center coding filters of size $7 \times 7$ and standard deviations of 1 and 2 pixels. To preprocess the CIFAR-10 dataset, we use a patch-based whitening method (filters of size $9 \times 9$, stride of 2, trained on $10^6$ samples). Hyperparameters of the CSNN are described in Table 1. Note that the number of epochs is proportional to the number of filters. Hyperparameters of the classification layer are described in Table 3. Gridsearch-optimized values are presented in Table 4 for SSTDP without weight normalization, in Table 4 for SSTDP with weight normalization, in Table 6 for S2-STDP without weight normalization, in Table 7 for S2-STDP with weight normalization, and in Table 8 for S2-STDP+PCN with weight normalization. In both feature extraction and classification layers, weights are initialized following a normal distribution $N(0.5, 0.01)$ and are clipped in $[0, 1]$ after each update. For more details about the hyperparameters, such as the R-STDP hyperparameters or the gridsearch ranges, please refer to the JSON files located in the `scripts/config/` folder of our source code.

Table 1: Hyperparameters of the CSNN trained with STDP.

| Hyperparameter | Value |
|---|---|
| epochs | 25 / 50 / 100 |
| number of filters | 16 / 64 / 128 |
| filter size | $5 \times 5$ |
| stride | 1 |
| padding | 0 |
| objective time ($t_{\text{target}}$) | See Table 2 |
| firing threshold ($V_{\text{th}}$) | See Table 2 |
| minimum threshold ($\text{Th}_{\text{min}}$) | 2 |
| threshold learning rate ($\eta_{\text{th}}$) | 1 |
| STDP saturation factor ($\beta$) | 1 |
| STDP learning rates ($A^+, A^-$) | 0.1, -0.1 |
| annealing | 0.95 |
| max-pool filter size | $4 \times 4$ |
| max-pool stride | 1 |
| max-pool padding | 0 |

Table 2: Task-dependent hyperparameters of the CSNN trained with STDP.

| Hyperparameter | MNIST | Fashion-MNIST | CIFAR-10 |
|---|---|---|---|
| objective time ($t_{\text{target}}$) | 0.75 | 0.80 | 0.95 |
| firing threshold ($V_{\text{th}}$) | 5 | 5 | 10 |

Table 3: Hyperparameters of the classification layer trained with SSTDP-based rules.

| Hyperparameter | Value |
|---|---|
| epochs | 100 |
| early stopping | 10 |
| annealing | 0.98 |
| STDP saturation factor ($\beta$) | 1 |
| firing threshold ($V_{\text{th}}$) | See Tables 4, 5, 6, 7, 8 |
| time gap ($g$) | See Tables 4, 5, 6, 7, 8 |
| STDP learning rates ($A^+, A^-$) | See Tables 4, 5, 6, 7, 8 |
| normalization factor ($w_{\text{norm}}$) | See Tables 5, 7, 8 |

Table 4: Hyperparameters of the classification layer optimized for SSTDP (without weight norm.) on CSNN-128.

| Hyperparameter | MNIST | Fashion-MNIST | CIFAR-10 |
|---|---|---|---|
| $V_{\text{th}}$ | 150 | 150 | 400 |
| $g$ | 0.07 | 0.05 | 0.0005 |
| $A^+$ | 0.05 | 0.025 | 0.075 |
| $A^-$ | -0.0 | -0.001 | -0.01 |

Table 5: Hyperparameters of the classification layer optimized for SSTDP (with weight norm.) on CSNN-128.

| Hyperparameter | MNIST | Fashion-MNIST | CIFAR-10 |
|---|---|---|---|
| $V_{\text{th}}$ | 200 | 350 | 200 |
| $w_{\text{norm}}$ | 0.2 | 0.2 | 0.2 |
| $g$ | 0.06 | 0.03 | 0.0005 |
| $A^+$ | 0.01 | 0.001 | 0.075 |
| $A^-$ | -0.05 | -0.2 | -0.075 |

Table 6: Hyperparameters of the classification layer optimized for S2-STDP (without weight norm.) on CSNN-128.

| Hyperparameter | MNIST | Fashion-MNIST | CIFAR-10 |
|---|---|---|---|
| $V_{\text{th}}$ | 300 | 350 | 550 |
| $g$ | 0.0075 | 0.025 | 0.0025 |
| $A^+$ | 0.05 | 0.1 | 0.1 |
| $A^-$ | -0.005 | -0.005 | -0.01 |

Table 7: Hyperparameters of the classification layer optimized for S2-STDP (with weight norm.) on CSNN-128.

| Hyperparameter | MNIST | Fashion-MNIST | CIFAR-10 |
|---|---|---|---|
| $V_{\text{th}}$ | 250 | 300 | 500 |
| $w_{\text{norm}}$ | 0.3 | 0.3 | 0.3 |
| $g$ | 0.03 | 0.02 | 0.005 |
| $A^+$ | 0.005 | 0.005 | 0.001 |
| $A^-$ | -0.1 | -0.075 | -0.075 |

Table 8: Hyperparameters of the classification layer optimized for S2-STDP+PCN (with weight norm.) on CSNN-128.

| Hyperparameter | MNIST | Fashion-MNIST | CIFAR-10 |
|---|---|---|---|
| $V_{\text{th}}$ | 250 | 350 | 450 |
| $w_{\text{norm}}$ | 0.3 | 0.3 | 0.3 |
| $g$ | 0.02 | 0.005 | 0.001 |
| $A^+$ | 0.001 | 0.0075 | 0.05 |
| $A^-$ | -0.1 | -0.2 | -0.2 |

## SuppM-III    Additional experiments

### SuppM-III.1    S2-STDP addresses the issues of SSTDP

In the paper, we demonstrate on the MNIST dataset that S2-STDP successfully addresses the issues of SSTDP regarding the limited number of STDP updates per epoch and the saturation of firing timestamps toward the maximum firing time. In this section, we provide an additional experiment with similar results on the Fashion-MNIST dataset. Figure 1 shows that, at epoch 30, S2-STDP increases the update ratio from 16% to nearly 100%, reduces the average firing time from 0.99 to 0.89, and augments its standard deviation from 0.008 to 0.05. As a result, our proposed methods based on S2-STDP enable training convergence at higher accuracies. It is important to mention that the resolution of these issues is not attributed to the additional weight normalization mechanism employed

by our methods. In Figure 2, we show on MNIST that SSTDP and S2-STDP exhibit similar behaviors with and without weight normalization, regarding update ratio, average firing time, and training accuracy.

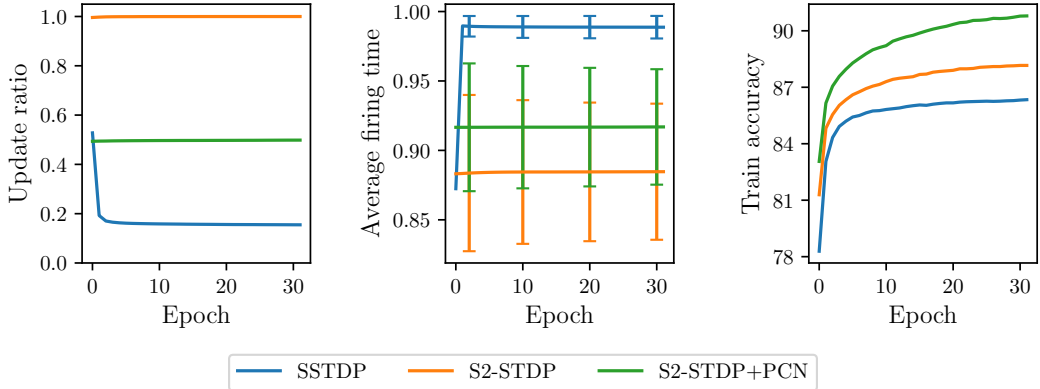

Figure 1: Update ratio, average firing time, and train accuracy per epoch in the classification layer trained on Fashion-MNIST. Our methods using S2-STDP significantly increase the number of updates per epoch and reduce the saturation of firing timestamps toward the maximum firing time. As a result, they enable training convergence at higher accuracies compared to SSTDP.

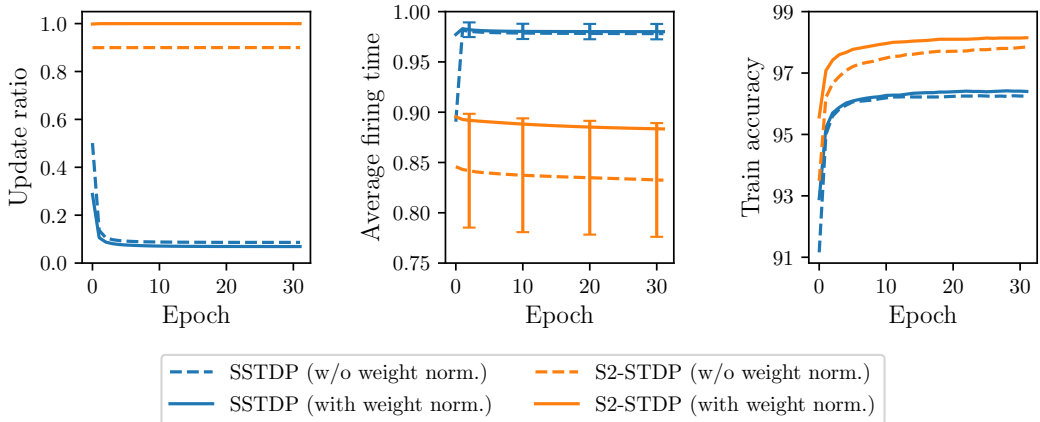

Figure 2: Update ratio, average firing time, and train accuracy per epoch in the classification layer trained on MNIST. S2-STDP successfully addresses the issues of SSTDP, with and without weight normalization.

S2-STDP pushes neurons to fire closer to the average firing time, and hence, to each other, compared to SSTDP. While this behavior may seem undesirable to improve class separability, we show in the main paper that ensuring accurate control over the output firing timestamps is more important than maximizing the firing time difference between the target and non-target neurons. Figure 3 provides, for each class of the MNIST test set, the distribution of firing time differences between the first non-target neuron to fire and the target neuron, in the classification layer trained with SSTDP and S2-STDP. These distributions show that the firing time differences of S2-STDP tend to be significantly smaller compared to SSTDP but result in fewer misclassified samples.

## SuppM-III.2    PCN enables neuron specialization

This work introduces the PCN architecture to improve the performance of our classification layer trained with S2-STDP.

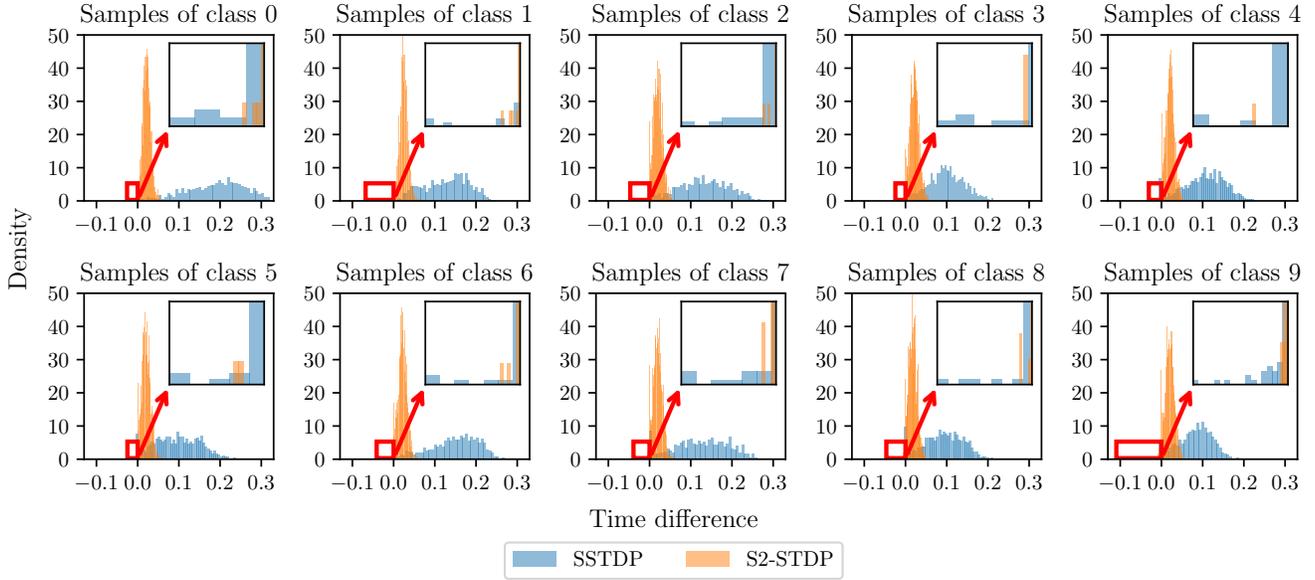

Figure 3: Distribution of firing time differences between the first non-target neuron to fire and the target neuron, in the classification layer on MNIST test samples. A negative time difference indicates that the target neuron fired after a non-target neuron, leading to misclassification of the sample. S2-STDP achieves a tighter distribution of time differences but results in fewer misclassified samples compared to SSTDP.

PCN represents each class by paired neurons and promotes specialization, such as one neuron learns to fire at the target desired timestamp and the other learns to fire at the non-target desired timestamp. The difference between the target and non-target firing timestamps is measured by the time gap $g$. Here, we analyze the actual firing time difference between paired neurons after training on the MNIST dataset. Figure 4 shows, for each class, the distribution of absolute firing time differences on test samples. As evidenced by the Gaussian shape distributions, paired neurons fire, on average, with a temporal interval of $g$. Hence, it provides evidence that PCN successfully enables neurons to specialize toward one of the two desired timestamps.

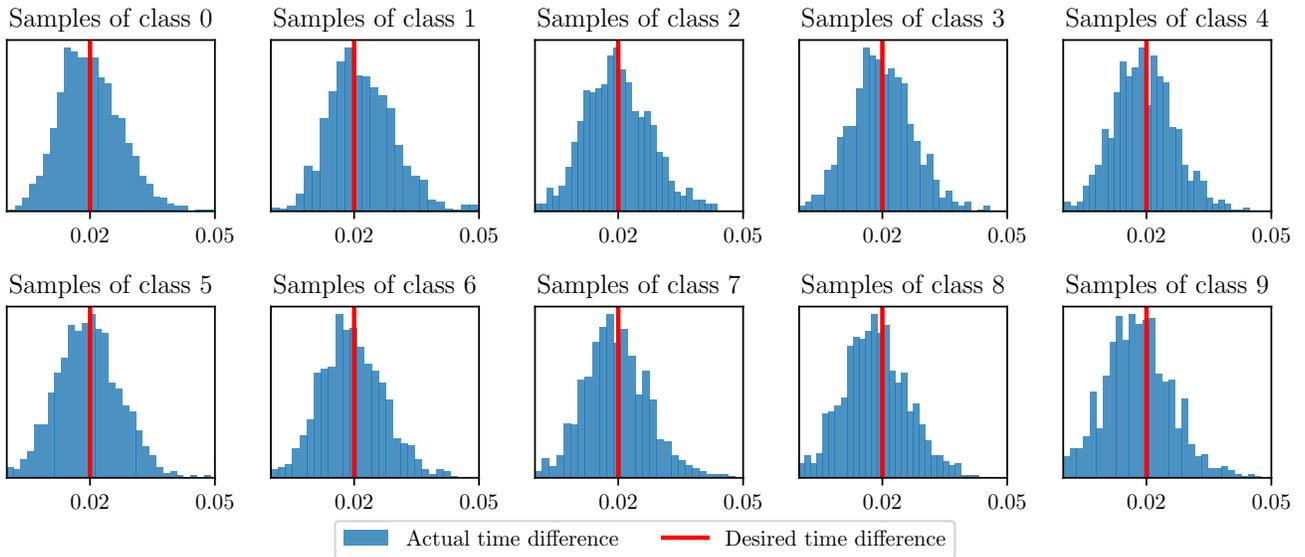

Figure 4: Distribution of absolute firing time differences between paired neurons on MNIST test samples, in the classification layer trained with S2-STDP+PCN.

In the paper, we provide evidence that neurons trained with S2-STDP in a PCN architecture are better at reaching the desired timestamps, especially for non-target samples. In this experiment, we show additional results on MNIST test samples to demonstrate this behavior. Figure 5 illustrates the distribution of time differences between neuron firing and desired timestamps, in the classification layer trained with S2-STDP and S2-STDP+PCN. As observed, S2-STDP+PCN achieves a significantly narrower distribution for non-target neurons, indicating that they fire, on average, closer to their desired timestamps. The distribution for target neurons is also slightly tightened, although to a lower extent. Nonetheless, the proportion of higher positive time differences (around 0.02), corresponding to target neurons firing after their desired timestamp, is reduced. It suggests that target neurons tend to fire farther away from non-target neurons, which may improve class separability.

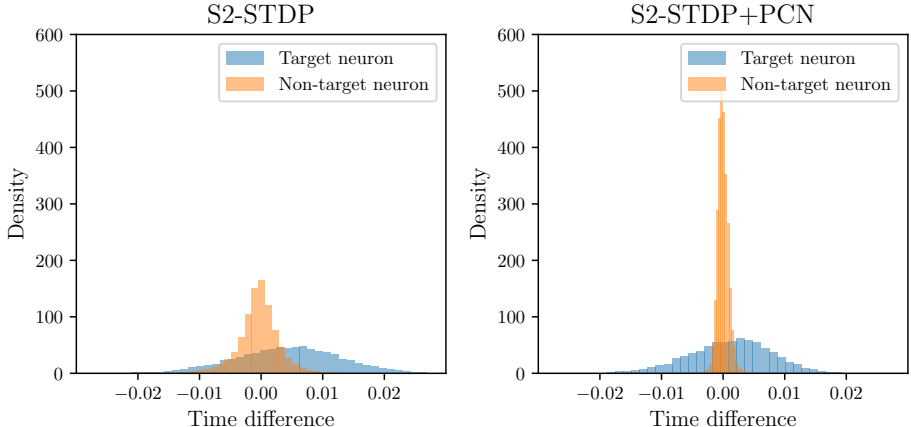

Figure 5: Distribution of time differences between neuron firing and desired timestamps, in the classification layer on MNIST test samples. Integrating a PCN architecture significantly tightens the non-target distribution, which indicates that neurons specializing toward non-target samples fire closer to their desired timestamps.

### SuppM-III.3 Effect of weight normalization on accuracy

Training with S2-STDP involves weight normalization to ensure that neurons maintain a similar weight average during training, and hence, equal chances of activation. In this section, we evaluate the effect of weight normalization on accuracy. Figure 6 compares the accuracy of SSTDP and S2-STDP, with and without weight normalization. The hyperparameters have been optimized independently for each classification layer model. Also, note that the original SSTDP training method does not employ weight normalization. We observe that weight normalization consistently improves the accuracy of S2-STDP across all datasets. Conversely, with SSTDP, weight normalization is beneficial only on MNIST and CIFAR-10. Without normalization, S2-STDP and SSTDP achieve comparable performance on Fashion-MNIST and CIFAR-10. However, our proposed training method comprising S2-STDP and weight normalization consistently outperforms SSTDP, both with and without weight normalization. More importantly, S2-STDP leverages compatibility with the PCN architecture, which further improves the accuracy of S2-STDP across all datasets, and without requiring any additional hyperparameters.

### SuppM-III.4 Robustness against the hyperparameter set

In the main paper, we show on the Fashion-MNIST dataset that PCN can effectively improve the accuracy of S2-STDP irrespective of the hyperparameters used. Figures 7 and 8 provide additional evidence on the MNIST and CIFAR-10 datasets, respectively.

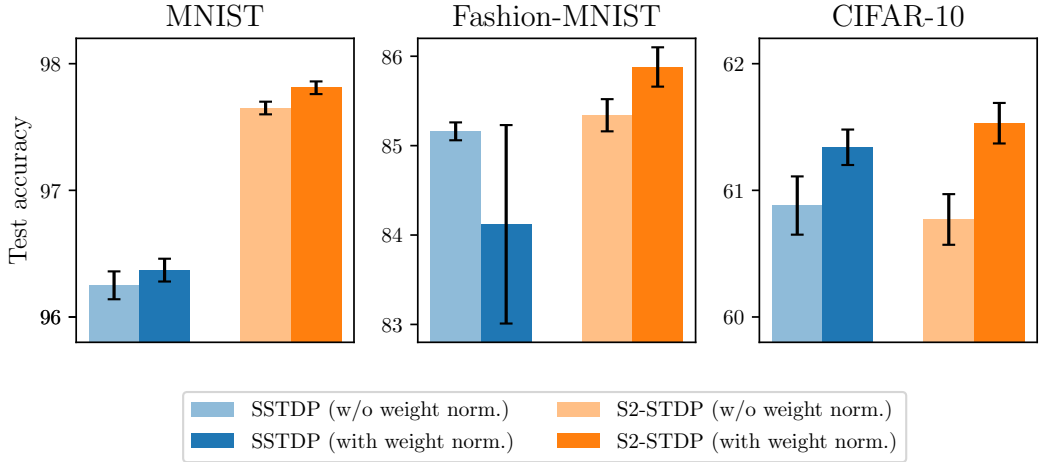

Figure 6: Accuracy of SSTDP and S2-STDP, both with and without weight normalization. Weight normalization consistently improves S2-STDP accuracy. As a result, our proposed training method comprising S2-STDP and weight normalization achieves higher accuracies compared to SSTDP.

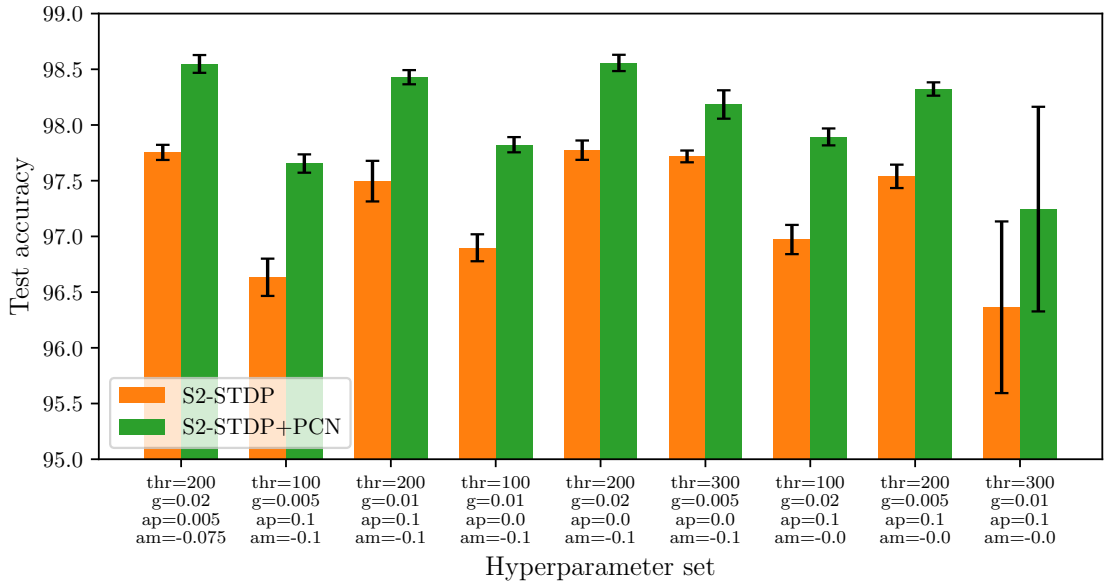

Figure 7: Accuracy of S2-STDP, with and without PCN, across different hyperparameter sets on MNIST. PCN always improves S2-STDP performance, without introducing any additional hyperparameters.



\bibliographystyle{Frontiers-Harvard}
\bibliography{references.bib}

\end{document}